\RequirePackage{fix-cm}
\documentclass[twocolumn,twoside]{svjour3}          % twocolumn
\usepackage{graphicx}
\usepackage{makecell, tabularx}
\usepackage[utf8]{inputenc} % allow utf-8 input
\usepackage[T1]{fontenc}    % use 8-bit T1 fonts
\usepackage{url}            % simple URL typesetting
\usepackage{booktabs}       % professional-quality tables
\usepackage{multirow,array}
\usepackage{amsfonts}       % blackboard math symbols
\usepackage{nicefrac}       % compact symbols for 1/2, etc.
\usepackage{microtype}      % microtypography
\usepackage{graphicx}
\usepackage{lipsum}
\usepackage{mathtools} 
\usepackage{amsmath}
\usepackage{svg}
\usepackage{extarrows}
\usepackage[]{threeparttable}
\usepackage{xcolor}
\usepackage[misc]{ifsym}
\usepackage{amssymb}% http://ctan.org/pkg/amssymb
\usepackage{pifont}% http://ctan.org/pkg/pifont
\usepackage{subcaption}
\usepackage[switch]{lineno}
\usepackage[colorlinks,linkcolor=red,anchorcolor=blue,citecolor=blue]{hyperref}
\usepackage[]{svg}
\usepackage{lmodern}% http://ctan.org/pkg/lm
\newcommand{\eg}{\textit{e}.\textit{g}.}
\newcommand{\etal}{\textit{et al}.}
\newcommand{\ie}{\textit{i}.\textit{e}.}

 \journalname{International Journal of Computer Vision}
\begin{document}

\title{A Survey of Representation Learning, Optimization Strategies, and Applications for Omnidirectional Vision}
% : A Survey and Benchmark}
% \title{Deep Learning for Omnidirectional Vision: A Survey and Benchmark}

\author{{Hao Ai} \textsuperscript{1}        \and
        {Zidong Cao} \textsuperscript{1} \and
        {Lin Wang} \textsuperscript{1,2,~\Letter} \\
        % \email{hai033@connect.hkust-gz.edu.cn, caozidong1996@gmail.com, linwang@ust.hk}
}

% \institute{%
% 	\begin{itemize}
% 		\item[\textsuperscript{\Letter}] L. Wang \\
% 		linwang@ust.hk
% 		\at
% 		\item[\textsuperscript{1}] Artificial Intelligence Thrust, HKUST GZ, Guangzhou, China
% 		\item[\textsuperscript{2}] Dept. of Computer Science and Engineering, HKUST, Hong Kong SAR, China
% 	\end{itemize}
% }

\maketitle
\vspace{-15pt}
\begin{abstract}
Omnidirectional image (ODI) data is captured with a field-of-view of $360^\circ \times 180^\circ$, which is much wider than the pinhole cameras and captures richer surrounding environment details than the conventional perspective images. In recent years, the availability of customer-level $360^\circ$ cameras has made omnidirectional vision more popular, and the advance of deep learning (DL) has significantly sparked its research and applications. This paper presents a systematic and comprehensive review and analysis of the recent progress of DL for omnidirectional vision. It delineates the distinct challenges and complexities encountered in applying DL to omnidirectional images as opposed to traditional perspective imagery. Our work covers four main contents: (i) A thorough introduction to the principles of omnidirectional imaging and commonly explored projections of ODI; (ii) A methodical review of varied representation learning approaches tailored for ODI; (iii) An in-depth investigation of optimization strategies specific to omnidirectional vision; (iv) A structural and hierarchical taxonomy of the DL methods for the representative omnidirectional vision tasks, from visual enhancement (\eg, image generation and super-resolution) to 3D geometry and motion estimation (\eg, depth and optical flow estimation), alongside the discussions on emergent research directions; (v) An overview of cutting-edge applications (\eg, autonomous driving and virtual reality), coupled with a critical discussion on prevailing challenges and open questions, to trigger more research in the community.

\end{abstract}

\begin{figure}[t]
    \centering        
    \includegraphics[width=1\linewidth]{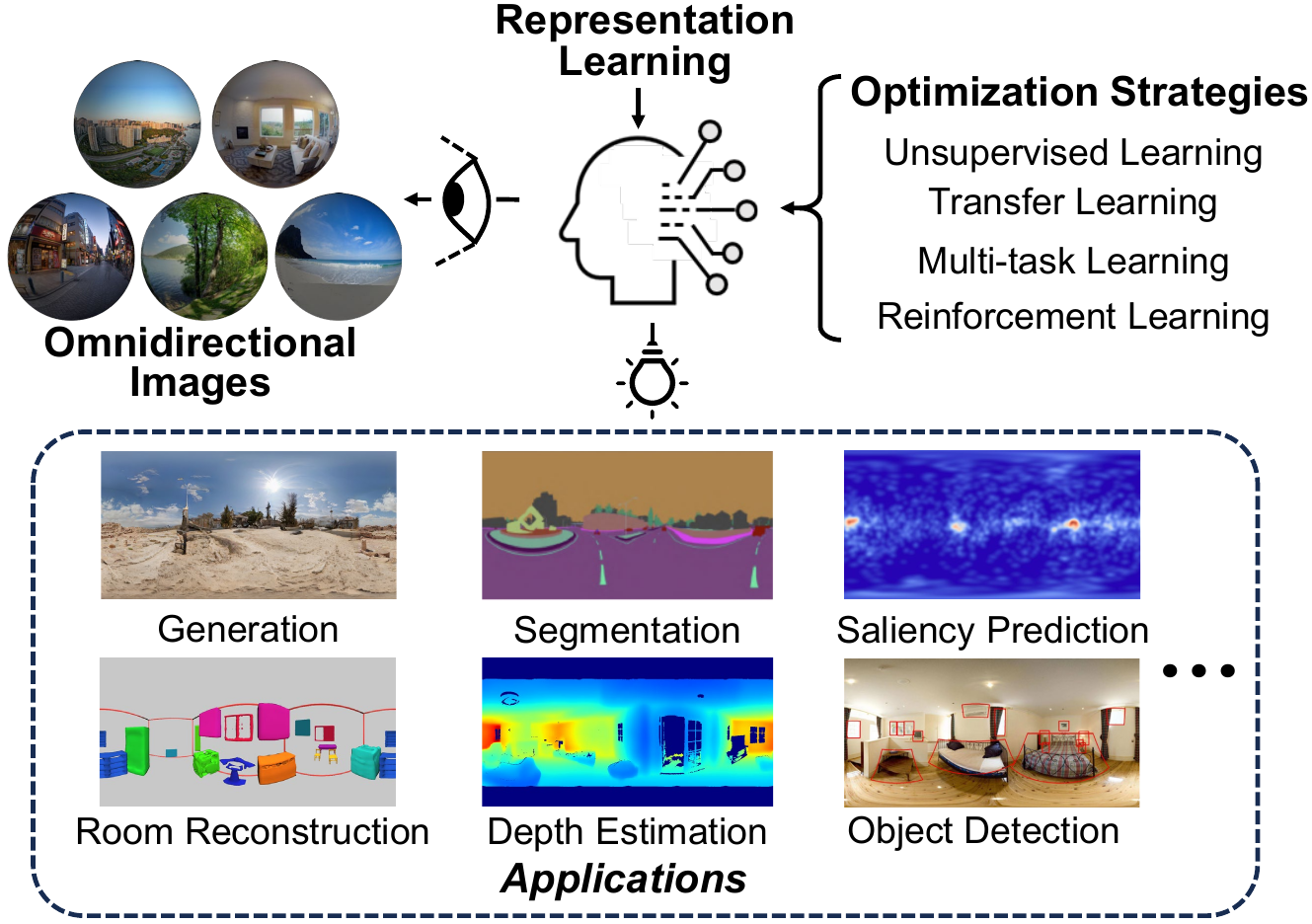}
    \vspace{-6pt}
    \caption{Overview of representation learning, optimization strategies, and applications for omnidirectional vision.}
    \label{fig:teaser}
    \vspace{-15pt}
\end{figure}

\vspace{-10pt}
\section{Introduction}
\vspace{-10pt}
\label{sec1}
\begin{figure*}[t]
    \centering        
    \includegraphics[width=1\textwidth]{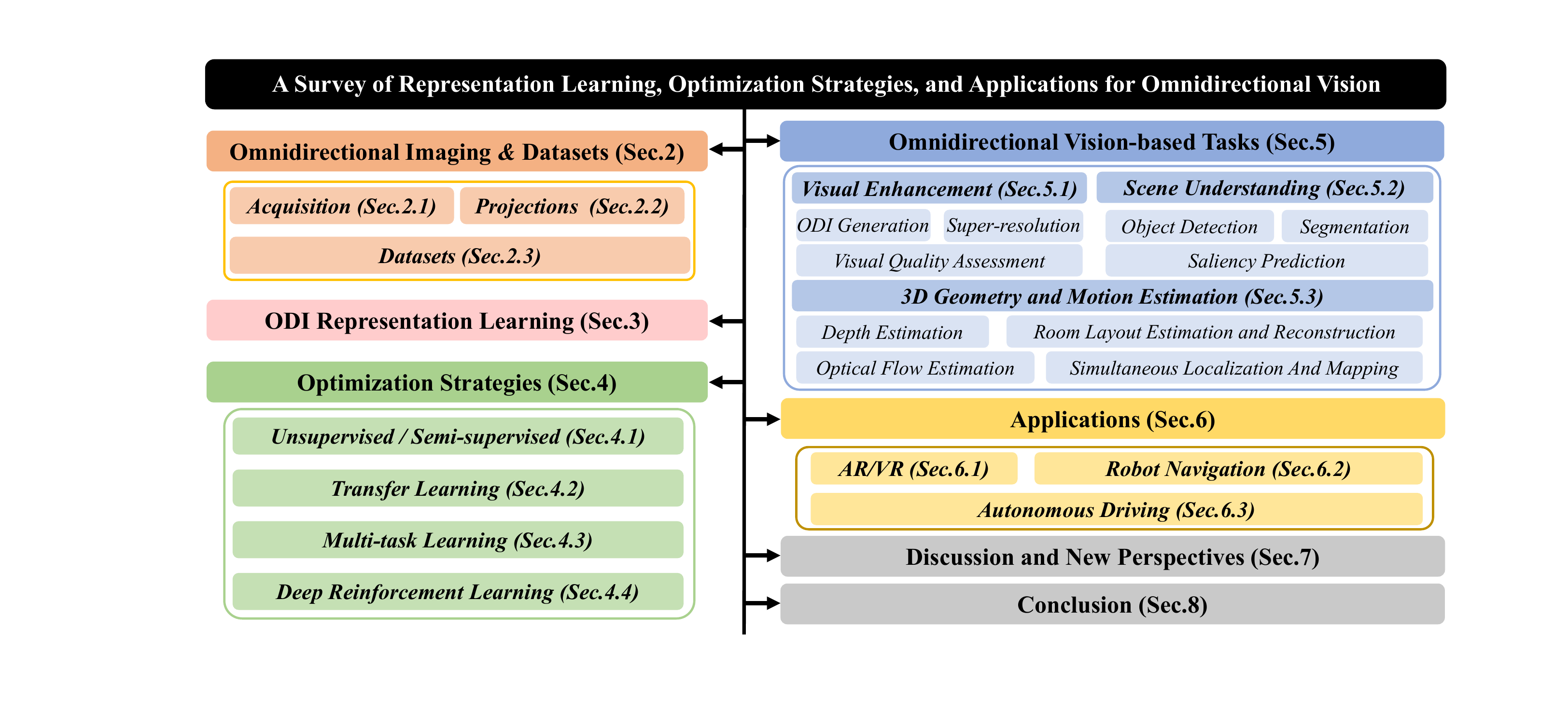}
     \vspace{-5pt}
 \caption{Hierarchical and structural taxonomy of omnidirectional vision with deep learning.}
    \label{fig:survey}
    \vspace{-15pt}
\end{figure*}
\begin{figure}[t!]
\centering
\includegraphics[width=0.65\linewidth]{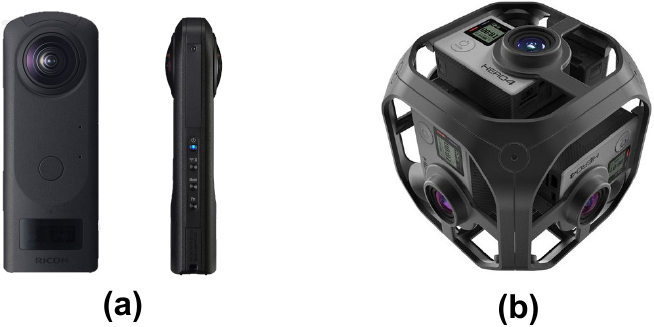}
    \vspace{-8pt}   
    \caption{Examples of $360^\circ$ cameras: (a) RICOH Theta Z1, and (b) GoPro Omni.}
    \label{fig:camera}
    \vspace{-15pt}
\end{figure}
With the rapid development of 3D technology and the pursuit of realistic visual experience, research interest in computer vision has gradually shifted from traditional 2D perspective images to omnidirectional images (ODIs)~\cite{Peleg1999StereoPW}, also known as the 360$^\circ$ images, panoramas\footnote{In this survey, panorama specifically refers to the equirectangular projection (ERP) format ODI.}, or spherical images. ODIs captured by the $360^\circ$ cameras yield a $360^\circ \times 180^\circ$ field-of-view (FoV), which is much wider than the pinhole cameras (maximum $180^\circ \times 180^\circ$) and fisheye cameras~\cite{Meuleman2021RealTimeSS} (about $220^\circ \times 180^\circ$); therefore, it can capture the entire surrounding environment by reflecting richer spatial information than the conventional perspective images and fisheye images. Due to the complete view and immersive viewing experience, ODIs have been widely applied to numerous applications, \eg, augmented reality (AR)$/$virtual reality (VR), autonomous driving, and robot navigation. As raw ODIs are captured on the sphere (Fig.~\ref{fig:teaser} Left), the two-dimensional (2D) projection formats of ODIs are required for coding, compression, transmitting, and analysis. Common 2D projection formats include Equirectangular projection (ERP), Cubemap projection (CP), and so on~\cite{Chen2018RecentAI},~\cite{HaiUyen2020SubjectiveAO}. As a novel data domain, ODIs have both domain-unique advantages (ultra-wide FoV of spherical imaging, comprehensive structural information of surrounding environments, multiple projection formats with distinctive advantages) and challenges (the demand for large receptive fields, robust rotation equivariance for spherical rotation, undeniable flaws in each projection format). These render the research on omnidirectional vision valuable yet challenging.

Recently, popular customer-level $360^\circ$ cameras (See Fig.~\ref{fig:camera}) have made omnidirectional vision more attractive, and the advance of deep learning (DL) has significantly promoted its research and applications, (See Fig.~\ref{fig:teaser}). In particular, the continual release of public datasets, \eg, Stanford2D3D~\cite{armeni2017joint}, Matterport3D~\cite{chang2017matterport3d} and PanoContext~\cite{Zhang2014PanoContextAW}, have rapidly enabled the data-driven DL methods to accomplish remarkable breakthroughs and often achieve the state-of-the-art (SoTA) performances on various vision tasks. Moreover, various deep neural network (DNN) models have been developed based on diverse architectures, ranging from convolutional neural networks (CNNs)~\cite{He2016DeepRL}, recurrent neural networks (RNNs)~\cite{medsker2001recurrent}, to vision transformers (ViTs)~\cite{dosovitskiy2020image}. In general, SoTA DL methods for ODI focus on three major aspects: \textbf{(i)} learning effective representations from diverse projections, \textbf{(ii)} novel learning strategies to utilize powerful knowledge of perspective images, \textbf{(iii)} customizing task-specificnetwork architectures.

In the past decade, several survey papers and reviews have been published in the omnidirectional vision community, including works by~\cite{Zou2021ManhattanRL},~\cite{daSilveira20223DSG},~\cite{Yu2023ApplicationsOD},~\cite{jiang20233d},~\cite{Zink2019Scalable3V},~\cite{Xu2020StateoftheArtI3},~\cite{yaqoob2020survey}, ~\cite{gao2022review}. Zink~\etal~\cite{Zink2019Scalable3V} and Yaqoob~\etal~\cite{yaqoob2020survey} have focused on techniques for 360$^\circ$ video streaming systems, particularly on transmission and compression. Similarly, Xu~\etal~\cite{Xu2020StateoftheArtI3} concentrated on the transmission, compression, and quality assessment of ODIs and omnidirectional videos (ODVs). Specializing on room layout estimation based on indoor ODIs, Zou~\etal~\cite{Zou2021ManhattanRL} reviewed and evaluated previous methods on numerous benchmark datasets, analyzing the design differences among various approaches. Subsequently, Gao~\etal~\cite{gao2022review} introduced the details of omnidirectional imaging systems and related optical technologies. With an emphasis on robotic applications, Gao~\etal introduced and summarized various scene understanding tasks based on ODIs, such as depth estimation, semantic segmentation, and simultaneous localization and mapping (SLAM). By contrast, Recent surveys~\cite{Zou2021ManhattanRL,daSilveira20223DSG,Yu2023ApplicationsOD,jiang20233d} collect and analyse approaches designed for 3D vision tasks, covering fields such as stereo matching, 3D reconstruction, and so on. Despite the extensive focus on how DL-based methods can effectively process ODIs or ODVs, there is a relative lack of exploration into the broader impact of DL's development on omnidirectional vision.

\begin{figure}[t!]
\centering
\includegraphics[width=0.75\linewidth]{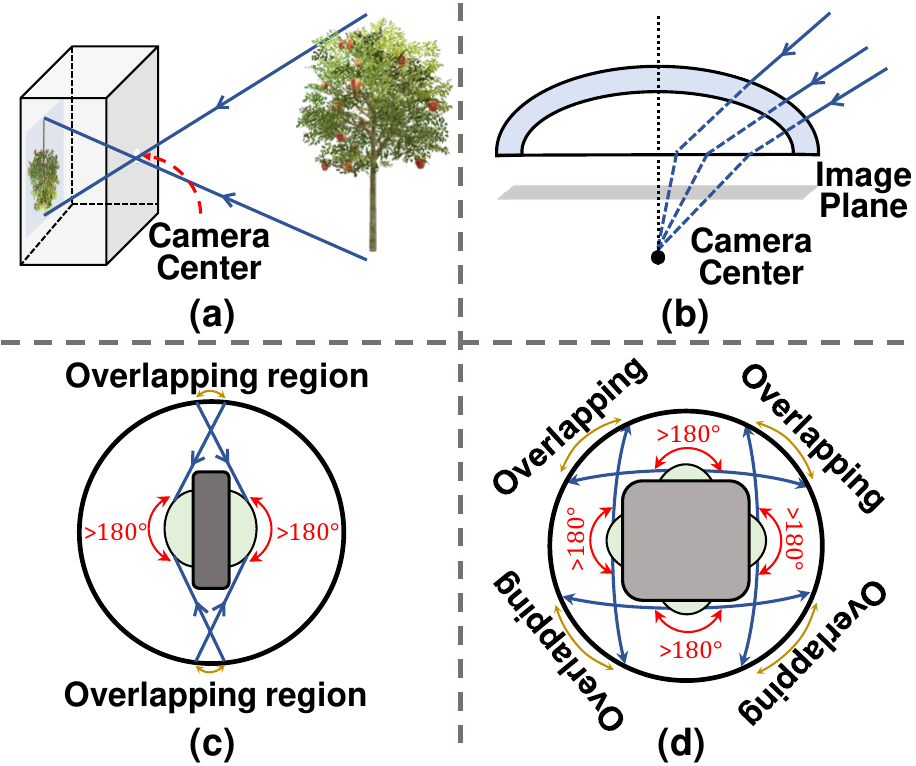}
\vspace{-5pt}
\caption{Imaging principles of several cameras: (a) Pinhole camera; (b) Fisheye camera; (c) 360$^\circ$ camera (dual-fisheye); (d) 360$^\circ$ camera (multi-fisheye).}
\vspace{-15pt}
\label{fig:imaging_prin}
\end{figure}

It can be observed that existing surveys mainly focus on specific domains. However, with the advent of deep learning, its impact on the entire omnidirectional vision community has been substantial. Systematically synthesizing, analyzing, and summarizing deep learning methods across a wide range of vision tasks is highly valuable. Therefore, compared to previous surveys, we provide a more thorough review, investigating how to collect and process ODIs for deep learning (Sec.~\ref{sec2}), how to learn representations (Sec.~\ref{sec3}) from ODIs, and how to optimize the models (Sec.~\ref{sec4}) for omnidirectional vision. Furthermore, while we review DL methods for appealing vision tasks in Sec.~\ref{sec5}, we go beyond simply compiling and summarizing algorithms; we critically analyze their foundations to categorize them and scrutinize the differences between methods tailored for ODIs versus those for perspective images. Fig.~\ref{fig:survey} shows the structural and hierarchical taxonomy of this study.

In summary, the major contributions of this study can be summarized as follows: (\textbf{\uppercase\expandafter{\romannumeral1}}) To the best of our knowledge, this is the \textbf{first} survey to comprehensively review and analyze the development of DL techniques for omnidirectional vision, including the omnidirectional imaging principle, diverse projection formats, datasets, representation learning, optimization strategies, a taxonomy, and applications, especially highlighting the differences and difficulties compared with 2D perspective images. (\textbf{\uppercase\expandafter{\romannumeral2}}) We summarize, if not all but representative, published top-tier conference/journal works (over 200 papers) in the last five years and conduct an analytical study of recent trends of DL on omnidirectional vision, both hierarchically and structurally. Moreover, we offer insights into the discussions for popular directions. (\textbf{\uppercase\expandafter{\romannumeral3}}) We have not only summarized representative methods and their key strategies for some popular omnidirectional vision tasks,~\eg, Fig.~\ref{fig:generation}, Fig.~\ref{fig:SR works}, Fig.~\ref{fig:VQA}, Fig.~\ref{fig:saliency}, Fig.~\ref{fig:Monocular depth estimation}, Fig~\ref{fig:Room Layout Estimation}, but also presented some representative methods' quantitative and qualitative results on benchmark datasets for better intra-task comparisons,~\eg, Tab.~\ref{tab:representation_comparison}, Tab.~\ref{tab:vqa/comparison-vqa}, Tab.~\ref{tab:segmentation/comparison-segmentation-sup}, 
 Tab.~\ref{tab:deph/cuboid-room-layout}, Tab.~\ref{tab:deph/comparison-depth}. (\textbf{\uppercase\expandafter{\romannumeral4}}) We provide insightful discussions of the practical applications and challenges yet to be solved and propose the potential of future directions to spur more in-depth research by the community. (\textbf{\uppercase\expandafter{\romannumeral5}}) We create an open-source repository that provides a taxonomy of all the mentioned works and code links. We will keep updating our open-source repository with new works in this area and hope it can shed light on future research. The repository link is \url{https://github.com/vlislab22/360_Survey}.

\vspace{-15pt}
\section{Omnidirectional Imaging and Projection}
\label{sec2}

\vspace{-3pt}
\subsection{Acquisition}
\vspace{-3pt}
\label{sec2.1}

As shown in Fig.~\ref{fig:imaging_prin}(a) and (b), a normal pinhole camera has a FoV less than 180$^{\circ}$ and thus captures a view at most a hemisphere, while a single fisheye camera usually has a wide FoV but less than 360$^\circ$, capturing a panoramic hemispherical image. In contrast, an ideal 360$^\circ$ camera can capture lights falling on the focal point from all directions, making the projection plane a whole spherical surface. According to the number of lenses, 360$^\circ$ cameras can be categorized into two types: (i) Cameras with back-to-back dual-fisheye lenses (Fig.~\ref{fig:imaging_prin}(c)); (ii) Cameras with more than two lenses (Fig.~\ref{fig:imaging_prin}(d). 

\begin{figure}[t!]
\centering
\includegraphics[width=\linewidth]{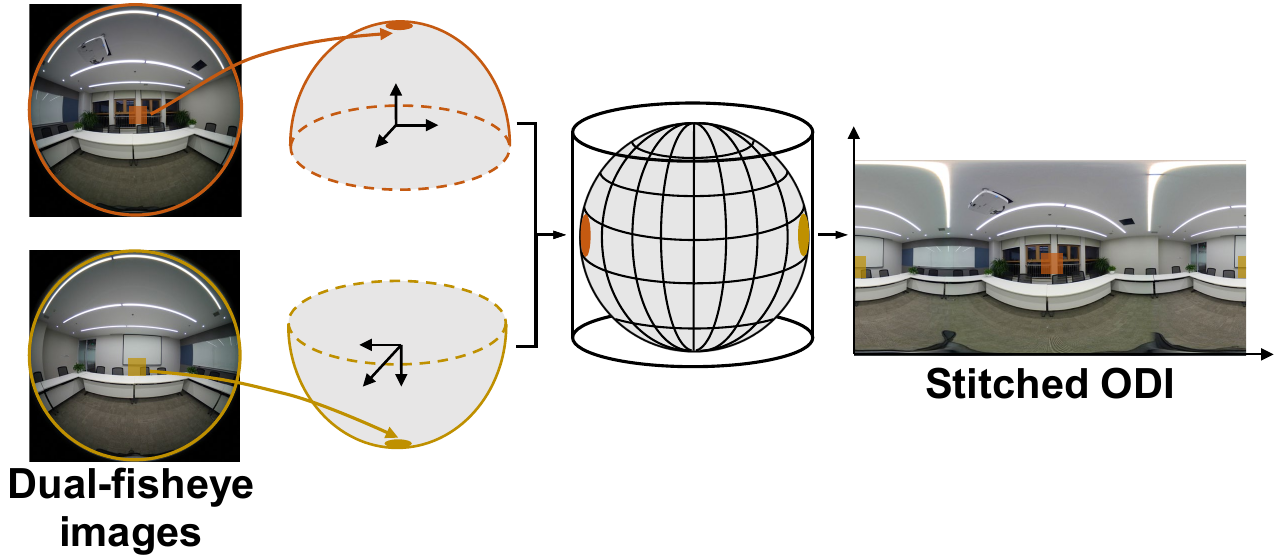}
    \vspace{-12pt}   
    \caption{Illustration of the process for stitching a pair of dual-fisheye images into an ERP format ODI.}
    \label{fig:stitching}
    \vspace{-15pt}
\end{figure}

The first type of 360$^\circ$ cameras requires a minimal number of lenses, making them cost-effective and user-friendly, which has led to widespread adoption in both industrial and consumer markets,~\eg, RICOH Theta\footnote{\url{https://www.ricoh360.com/theta}} (Fig.\ref{fig:camera}(a)). However, stitching two back-to-back fisheye images into a seamless omnidirectional image (ODI) presents significant challenges. These challenges arise primarily from (1) the displacement of the optical centers of the two fisheye cameras, and (2) limited information and severe distortions in the overlapping regions between the two cameras (Fig.~\ref{fig:imaging_prin}(c)). To address these challenges and generate high-quality ODIs, considerable efforts have been dedicated~\cite{lo2020photometric},~\cite{roberto2020using},~\cite{flores2024generating},~\cite{li2019attentive},~\cite{lo2018image},~\cite{lo2021efficient}. Typically, the stitching process, as illustrated in Fig.~\ref{fig:stitching}, involves three key steps: (1) transforming fisheye images into spherical coordinates, (2) aligning the spherical views within a unified spherical coordinate system, and (3) blending the aligned views while resolving inconsistencies to obtain the final ODI. For the second type, more lenses bring higher precision and less inconsistency for the final stitched results. This type of $360^{\circ}$ cameras are professional-level and expensive, \eg, GoPro Omni (six lenses)\footnote{\url{https://gopro.com/en/us/news/omni-is-here}} (Fig.~\ref{fig:camera}(b)). In addition, some studies~\cite{ullah2020automatic},~\cite{zia2019360} have mounted multiple fisheye cameras on the wings of drones, directly generating high-quality ODIs using the captured fisheye images and corresponding camera parameters. Meanwhile, stitching multiple fisheye images to produce ODIs has also been applied in SLAM systems~\cite{ji2020panoramic}. In this work, we place greater emphasis on the representation and applications of ODIs. To gain a better understanding of the imaging principles of ODIs, we recommend readers refer to~\cite{gao2022review},~\cite{Yan2023DeepLO}.

% In particular, Lo~\etal~\cite{lo2020photometric, lo2021efficient} and Flores~\etal~\cite{flores2024generating} introduced fisheye camera calibration techniques to project the fisheye image centers onto the poles of the spherical view. Furthermore, to address the color and intensity disparities caused by the positional displacement of the fisheye camera centers, Lo~\etal~\cite{lo2020photometric, lo2021efficient} proposed a photometric compensation loss to enhance consistency in overlapping regions. Notably, Li~\etal~\cite{li2019attentive} presented an end-to-end convolutional network that learns transformation rules for converting dual-fisheye images to ODIs, guided by a novel attentive quality assessment framework.}

\vspace{-15pt}
\subsection{Projections of ODI}
\vspace{-3pt}
\label{sec2.2}

\begin{figure}[t!]
\centering
\includegraphics[width=0.90\linewidth]{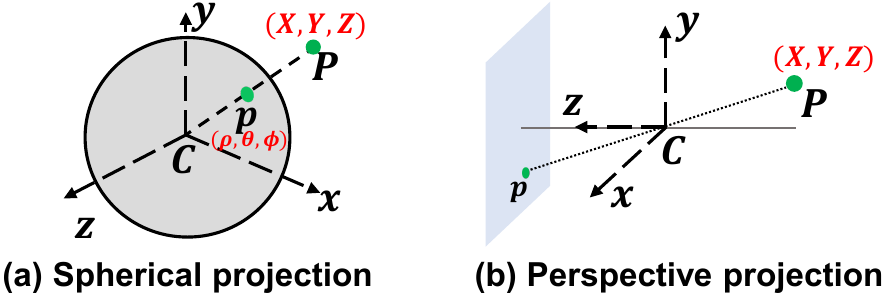}
    \vspace{-5pt}
\caption{Spherical projection \textit{v.s.} perspective projection.}
    \label{fig:sprojection}
    \vspace{-10pt}
\end{figure}

\begin{figure*}[t]
    \centering
    \includegraphics[width=0.92\textwidth]{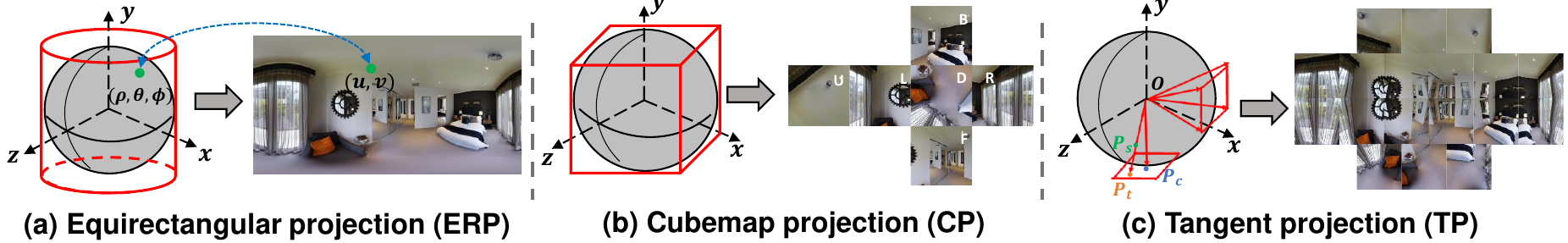}
        \vspace{-5pt}
    \caption{Illustration of several representative projections: (a) ERP; (b) CP; (c) TP.}
    \label{fig:imaging}
    \vspace{-10pt}
\end{figure*}

\noindent \textbf{Spherical Projection.} A 360$^\circ$ camera can be modeled as a camera center with a surrounding unit sphere, which projects all visible points in the surroundings onto the sphere surface. Therefore, for a 360$^\circ$ camera, intrinsic parameters such as focal length or distortion values are not required to consider~\cite{Krolla2014SphericalLF},~\cite{daSilveira20223DSG},~\cite{Li2005SphericalSF}. As shown in Fig.~\ref{fig:sprojection}(b), given a point $P= \left [x,y,z\right ]^{T}$ in space, the spherical projection of $P$ is the intersection $p$ of the sphere surface with the line $\overrightarrow{CP}$ joining point $P$ and the sphere center $C$. To obtain the spherical coordinate of $p$, we first define that a 360$^\circ$ camera coordinate system is set at the
center of a unit sphere with $f=1$ (radius), and the direction of $\overrightarrow{CP}$ is determined by $\theta$ (polar angle) and $\phi$ (azimuth angle). Accordingly, the spherical coordinate of $p$ can be represented as $p = \left [ \sin\theta\cos\phi, \sin\theta\sin\phi, \cos\theta \right ]^{T}$. Following the relationships between $P$ and $p$, we can obtain the spherical projection as follows:
\begin{equation}
    \begin{array}{|c|}
         \rho  \\
         \theta \\
         \phi \\
    \end{array} = 
    \begin{array}{|c|}
         (x^2+y^2+z^2)^{1/2} \\
         \arctan(x/z) \\
         \arccos(y/\rho) \\
    \end{array} \ , \ 
    \begin{array}{|c|}
         x  \\
         y \\
         z \\
    \end{array} = 
    \begin{array}{|c|}
         \rho\sin(\theta)\sin(\phi) \\
         \rho\cos(\phi) \\
         \rho\cos(\theta)\sin(\phi) \\
    \end{array}.
    \label{eq.1}
\end{equation}

\noindent \textbf{Equirectangular Projection (ERP).} In the industry, there lack the efficient image encoding and storage techniques for the images represented in the spherical space. Therefore, for processing and storing ODIs, the raw spherical data is always projected from the sphere to the 2D plane, as shown in Fig.~\ref{fig:imaging}(a). Equirectangular projection (\textbf{ERP}) is the most popular projection format, which uniformly samples grids from the spherical surface. The horizontal unit angle is $\varDelta\theta=2\pi/w$ and the vertical unit angle is $\varDelta\varphi=\pi/h$. In particular, if the horizontal and vertical unit angles are equal, the width $w$ is twice of height $h$. In a word, each pixel coordinate $(u,v)$ in ERP can be mapped to the spherical coordinate $(\theta,\phi)=(u \cdot \varDelta\theta,v\cdot \varDelta\varphi)$ and vice versa. It is notable that ERP introduces unavoidable shortcomings such as redundant samples and horizontal stretching, which can result in geometric distortions, especially in regions near the two poles~\cite{Kim20133DSR,Zhao2014SPHORBAF}.

\begin{figure}[t!]
\centering
\includegraphics[width=0.65\linewidth]{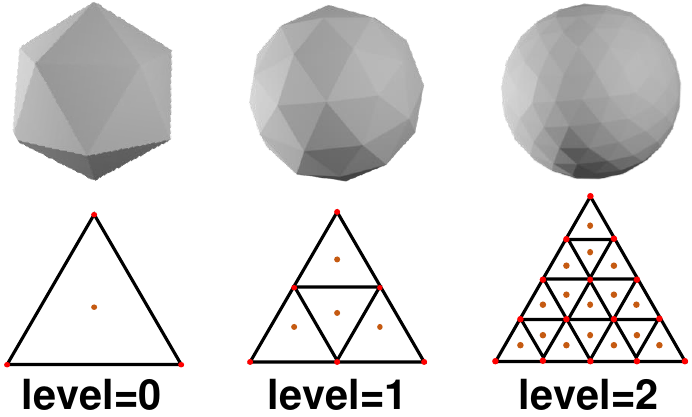}
 \vspace{-5pt}
\caption{\textbf{Top}: Icosahedron projection; \textbf{Bottom}: Subdivision of an icosahedron's face.}
    \label{fig:subdivision}
    \vspace{-15pt}
\end{figure}

\noindent \textbf{Cubemap Projection (CP).} To alleviate the distortion problem in ERP, cubemap projection (CP) is proposed to project the sphere surface to six cube faces with the FoV of $90^{\circ}\times90^{\circ}$. As depicted in Fig.~\ref{fig:imaging}(b), each face has the equal-side length $w$ and the distance between the face center and sphere center is $\frac{w}{2}$. The cube faces are denoted as $f_i$, $i \in \{B,D,F,L,R,U\}$, representing back, down, front, left, right, and up, respectively. By setting the cube center as the origin, the extrinsic matrix of each face can be simplified into $90^\circ$ or $180^\circ$ rotation and zero translation matrices~\cite{wang2020bifuse}. Given a plane pixel $f_i$, 
we transform $f_i$ to the front plane (identical to the Cartesian coordinates) and calculate $(\theta, \phi)$ via Eq.~\ref{eq.1}.

\noindent \textbf{Tangent Projection (TP).} To preserve the performance of perspective models on high-resolution spherical data,~\cite{eder2020tangent} proposes rendering the omnidirectional image as a set of local planar image grids tangent to the subdivided icosahedron, called tangent projection (\textbf{TP}). Specifically, TP is the gnomonic projection~\cite{o1962introduction}, a non-conformal projection from point $P_s$ on the sphere surface with the sphere center $O$ to point $P_t$ in a tangent plane with center $P_c$\footnote{\url{https://mathworld.wolfram.com/GnomonicProjection.html}}, as shown in Fig.~\ref{fig:imaging}(c). For a pixel on the ERP image $P_{e}(u_e,v_e)$, we first calculate its corresponding point $P_{s}(\theta=u_{e} \cdot \vartheta, \phi=v_{e} \cdot \varphi)$ on the unit sphere, following the transformation in ERP format. The projection from $P_{s}(\theta,\phi)$ to $P_t(u_t,v_t)$ is defined as:
\begin{equation}
\begin{aligned}
&    u_t=\frac{\cos(\phi)\sin(\theta-\theta_c)}{\cos{c}}, \\
&    v_t=\frac{\cos(\phi_c)\sin(\phi)-\sin(\phi_c)\cos(\phi)\cos(\theta-\theta_c)}{\cos(c)}, \\
&   \cos(c)=\sin(\phi_c)\sin(\phi)+\cos(\phi_c)\cos(\phi)\cos(\theta-\theta_c),
\end{aligned}
\label{eq.2}
\end{equation}

\noindent where $(\theta_c,\phi_c)$ is the spherical coordinate of the tangent plane center $P_c$, and $(u_t,v_t)$ is the intersection coordinate of the tangent plane and the extension line of $\overrightarrow{OP_s}$. The inverse transformations are formulated as:
\begin{equation}
\begin{split}
& \theta = \theta_c + \tan^{-1}(\frac{u_t\sin(c)}{\gamma\cos(\phi_c)\cos(c)-v_t\sin(\phi_c)\sin(c)}), \\
& \phi=\sin^{-1}(\cos(c)\sin(\phi_c)+\frac{1}{\gamma}v_t\sin(c)\cos(\phi_c)),
\end{split}
\label{eq.3}
\end{equation}
where $\gamma=\sqrt{u_{t}^{2}+v_{t}^{2}}$ and $c=\tan^{-1}\gamma$. With Eqs. \ref{eq.2} and \ref{eq.3}, we can build one-to-one forward and inverse mapping functions between the spherical coordinates and pixels on the tangent patches~\cite{li2022omnifusion}.

\noindent \textbf{Polyhedron Projection.} To mitigate the distortions from projecting the spherical images into the planar representations and maintain the continuity of spherical images, polyhedron projection (PP)~\cite{Lee2019SpherePHDAC} is proposed to approximate the spherical geometry based on successive divisions of spherical polyhedrons, \eg, octahedron projection, icosahedron projection. Specifically, each face in a spherical polyhedron can be subdivided into four smaller faces to achieve higher resolution and less distortion~\cite{Lee2019SpherePHDAC}. 
Using the icosahedron projection as an example, we can derive $20\times4^l$ faces for an icosahedron mesh at the default subdivision level $l$ (See Fig.~\ref{fig:subdivision}).

\noindent \textbf{Other projections.} Apart from the aforementioned popular projection formats, there are several other formats supported by the 360Lib software package for coding and processing~\cite{HaiUyen2020SubjectiveAO}. They include adjusted equal-area projection (AEP), truncated square pyramid projection (TSP), adjusted cubemap (ACP), rotated sphere projection (RSP), equatorial cylindrical projection (ECP), equiangular cubemap (EAC), and hybrid equiangular cubemap (HEC). Especially, as the map-based projection, AEP adaptively decreases the sampling rate in vertical coordinates and avoids the oversampling problem in ERP. Equi-Angular Cubemap (EAC) projection~\cite{Chen2018RecentAI} keeps spatial sampling rates for different sampling locations on the cube faces to alleviate the distortions in CP.

\begin{table*}[htbp]
\caption{The summary of ODI image and video datasets. $*$ indicates `not available', `Ref.' indicates `Reference image', and `Dist.' indicates `Distorted image'. For the purposes of datasets, \textbf{G}: Image Generation, \textbf{SR}: Image Super-resolution, \textbf{QA}: Image Quality Assessment, \textbf{OD}: Object Detection, \textbf{SS}: Semantic Segmentation, \textbf{DE}: Depth Estimation, \textbf{SP}: Saliency Prediction, \textbf{RL}: Room Layout Estimation and Reconstruction.}
\vspace{-5pt}
\centering
\resizebox{0.96\linewidth}{!}{ 
    \begin{tabular}{c|c|c|c|c}
    \toprule
    Task & Dataset & Data size& Resolution& Source\\ 
    \midrule
    \multirow{4}*{\textbf{G}}& 360SP~\cite{Chang2018Generating3O}&15,730 & 512$\times$ 256 & Real-world\\
    &HDR360-UHD~\cite{Chen2022Text2Light}& 4,392& 8192$\times$ 4096 & Real-world\\
    &Laval Indoor HDR~\cite{Gardner2017LearningTP}&2,100 & 2048$\times$ 1024 &Real-world\\
    &Laval Outdoor HDR~\cite{hold2019deep}& 205 & 2048$\times$ 1024 &Real-world\\ \hline
    \multirow{4}*{\textbf{SR}}&ODI-SR~\cite{Deng2021LAUNetLA} & 1,000 & 2408$\times$1024 &Real-world\\
    &Flickr360~\cite{Cao2023NTIRE2C} & 3,150 & 2408$\times$1024 & Real-world \\ 
    &ODV360 (Video)~\cite{Cao2023NTIRE2C} & 250 & 2160$\times$1080 & Real-world\\ \hline
    \multirow{5}*{\textbf{QA}}&Upenik~\cite{Upenik2016TestbedFS} & 6 (Ref.) \& 54 (Dist.) & 3000$\times$1500 &  Real-world  \\
    &CVIQD~\cite{Sun2017CVIQDSQ} & 16 (Ref.) \& 528 (Dist.) & 4096$\times$2048 & Real-world  \\
    & OIQA~\cite{Duan2018PerceptualQA} & 16 (Ref.) \& 320 (Dist.) & 13320$\times$6660 &  Real-world  \\
    &IQA-ODI~\cite{Yang2021SpatialAN} & 120 (Ref.) \& 960 (Dist.) & 7680$\times$3840 &   Real-world  \\
    &JUFE-VRIQA~\cite{Liu2024PerceptualQA} & 258 (Ref.) \& 1032 (Dist.) & 8192$\times$4096 & Real-world  \\\hline
     \multirow{7}*{\textbf{OD}}& 360-Indoor~\cite{chou2020360} &3,000&1920 $\times$960& Real-world  \\
     &ERA~\cite{yang2018object}&903&3840$\times$1920&Real-world  \\
     &OVS~\cite{yu2019grid}&600 & 2000$\times$1000&Real-world  \\
     &PANDORA~\cite{Xu2022PANDORAAP}&3,000 &1920$\times$960& Real-world  \\
     &VOC360~\cite{zhao2020spherical} &21,755&1024$\times$512&Synthetic\\
    &COCO-Men~\cite{zhao2020spherical} &7,000&1024$\times$512&Synthetic\\
    &FlyingCars~\cite{coors2018spherenet}&6000&512$\times$256&Synthetic\\ \hline
    \multirow{4}*{\textbf{SS}}&WildPASS~\cite{Yang2021CapturingOC}&2,500&2048$\times$400&Real-world\\
    &DensePASS~\cite{ma2021densepass}&100 &2048$\times$400& Real-world\\   &SynPASS~\cite{Zhang2022BehindED}&9,080&1024$\times$512& Synthetic\\
    & Omni-SYNTHIA ~\cite{zhang2019orientation}& 2,269 & 4192$\times$2096 & Synthetic\\ \hline
    \multirow{5}*{\textbf{DE}}&    Deep360~\cite{Li2022MODEMO}& 1,2000 (RGB) \& 3,000 (Depth)& 1024$\times$512 & Synthetic\\
    &Depth360~\cite{feng2022360} &30,000&1.03Mpx&Synthetic\\
    &Pano3D~\cite{Albanis2021Pano3DAH} &29,900&1024$\times$512& Synthetic\\
    &PanoSUNCG~\cite{wang2018self} &25k &1024$\times$512&Synthetic\\
    &360D~\cite{zioulis2018omnidepth} &35,977&1024$\times$512& Synthetic\\ \hline
    \multirow{5}*{\textbf{SP}}&Salient360!~\cite{rai2017dataset} & 85 & 18332$\times$9166 & Real-world\\
    &AOI~\cite{xu2021saliency} &600& 8000$\times$4000&   Real-world\\
    &Wild-360 (Video)~\cite{cheng2018cube} &85& 7680$\times$3840& Real-world\\
    &PVS-HM (Video)~\cite{xu2018predicting}&76&7680$\times$3840&Real-world\\
    &SVR~\cite{sitzmann2018saliency} &22&8192$\times$ 4096&  Synthetic\\ \hline
    \multirow{2}*{\textbf{RL}}&    ZInD~\cite{cruz2021zillow} &71,474&2048$\times$1024&  Synthetic\\
    &Pano3DLayout~\cite{pintore2021deep3dlayout}&107&1024$\times$512&Synthetic\\ \hline
    \textbf{SS}, \textbf{DE}&Gibson~\cite{xiazamirhe2018gibsonenv} &572 scenes&$-$& Synthetic \\ \hline
    \textbf{DE}, \textbf{RL}&Shanghaitech-Kujiale Indoor 360~\cite{jin2020geometric} &3,550&1024$\times$512& Synthetic \\ \hline
    \textbf{OD},\textbf{RL} & PanoContext~\cite{Zhang2014PanoContextAW} & 700 & 9104$\times$4552 & Real-world\\ \hline
    \textbf{G}, \textbf{SR}, \textbf{QA} & 
    SUN360*~\cite{xiao2012recognizing} &67,538 & 
    9104$\times$4552 & Real-world\\ \hline
    \textbf{DE}, \textbf{RL}, \textbf{SS} & Matterport3D~\cite{chang2017matterport3d} & 10,800 & 2048$\times$1024 & Real-world \\\hline
    \textbf{G}, \textbf{SR}, \textbf{SS}&360+$x$ (Video)~\cite{chen2024360+}  &28 scenes&5760$\times$2880& Real-world \\ \hline
    \textbf{OD}, \textbf{DE}, \textbf{RL}&Stanford2D3D~\cite{armeni2017joint} & 1,413 & 4096$\times$2048 & Real-world \\ \hline
    \textbf{SS}, \textbf{DE}, \textbf{RL} &Replica~\cite{straub2019replica} &18 scenes&$-$& Synthetic\\ \hline
     \textbf{OD}, \textbf{SS}, \textbf{DE}, \textbf{G}, \textbf{RL}&Structured3D~\cite{zheng2020structured3d} & 196,515 &1024$\times$512&Synthetic\\
    \bottomrule
\end{tabular}}
\vspace{-15pt}
\label{table:dataset}
\end{table*}

\vspace{-15pt}
\subsection{ODI datasets}
\label{sec2.3}
\vspace{-5pt}

As DL is data-driven, the performance of DL-based methods is intricately linked to the qualities and quantities of available data. In other words, ODI and omnidirectional video (ODV) datasets are critical for the development of DL methods on omnidirectional vision. Therefore, we have collected popular datasets for various tasks from the existing literature, listed in Tab.~\ref{table:dataset}. We provide information on the scale and maximum resolution of the datasets. Furthermore, we delineate whether each dataset originates from real-world sources. Notably, synthetic datasets are crafted using computer graphics and rendering technologies. We provide details of several widely used benchmark datasets below.
\begin{itemize}
    \item[$-$] SUN360~\cite{xiao2012recognizing}: A large-scale ODI dataset that contains 80 categories and 67,583 real-world ERP format ODIs. Notably, all ODIs are collected from the Internet\footnote{\url{www.360cities.net}}. 
    % Due to the large amount and high resolution (up to 9104 $\times$ 4552 pixels), SUN360 is widely used in low-level vision tasks, such as, generation, super-resolution, quality assessment. 
    Notably, despite its widespread use, the official download link is presently unavailable online. 
    \item[$-$]  Matterport3D~\cite{chang2017matterport3d}: A large-scale RGB-D ODI dataset comprising 90 building-scale scenes, encompassing 10,800 panoramas, depth maps, and associated annotations for surface reconstructions, camera poses, and 2D and 3D semantic segmentation labels. Notably, each panorama is constructed from 18 RGB-D images captured using a Matterport camera~\footnote{\url{https://matterport.com/}}, albeit with a limitation in coverage of the north and south poles. The dataset is split into 61 scenes for training, 11 for validation, and 18 for testing.
    \item[$-$] Stanford2D3D~\cite{armeni2017joint}: A RGB-D dataset consisting of 1,413 RGB panoramas from six real-world areas. Besides, the panoramas are coupled with corresponding depth maps, surface normal information, and semantic annotations. Each panorama is also stitched by 18 perspective images using the Matterport camera; therefore, Poles' color information is missing.
    \item[$-$] PanoContext~\cite{Zhang2014PanoContextAW}: This dataset has 700 panoramas of room environment from SUN360, including 418 bedrooms and 282 living rooms. The annotations of cuboid room layout and object bounding boxes are manually annotated by a self-designed WebGL annotation tool in the browser.
\end{itemize}
\vspace{-6pt}
Due to space constraints, we do not provide detailed descriptions for each dataset. However, we have included the source of each dataset, allowing researchers in the field to quickly understand the relevant datasets. Additionally, in our open-source repository, we have provided download links for each open dataset, making it easier for researchers to access the datasets they want.

\vspace{-15pt}
\section{ODI Representation Learning}

\label{sec3}
In this section, we mainly focus on how to extract powerful and general representations from ODIs, thus we will not discuss networks specifically designed for downstream vision tasks. In particular, based on the input projection format, we categorize these methods into: \textit{1) Euclidean space-based methods} for planar projections,~\eg, ERP, and \textit{2) Non-Euclidean space-based methods} for Non-Euclidean projections,~\eg, icosahedron meshes.
\begin{table}[!t]
    \centering
    \caption{State-of-the-art Results for spherical image classification on SPH-MNIST and SPH-CIFAR10.
    }
    \vspace{-5pt}
    \label{tab:representation_comparison}
    \resizebox{0.95\linewidth}{!}{ 
    \setlength{\tabcolsep}{4pt}
    \begin{tabular}{c|c|c|c}
    \toprule
    &Method & $\#$ Para.&Acc (\%) $\uparrow$\\
    \midrule
    \multirow{6}*{\shortstack{SPH-\\MNIST}}&VGG~\cite{simonyan2014very}+KTN~\cite{Su2019KernelTN}&294M&97.94\\
    &SphereNet~\cite{coors2018spherenet}&196k&94.41\\
    &SpherePHD~\cite{Lee2019SpherePHDAC}&57K&94.08\\
    &Spherical GCN~\cite{Yang2020RotationEG}&60k&94.42\\
    &PanoSwin-Tiny~\cite{ling2023panoswin}&30M&99.85\\
    &SPHTR~\cite{cho2024sampling}&60k&95.01\\
    \midrule
    \multirow{4}*{\shortstack{SPH-\\CIFAR10}}&SpherePHD~\cite{Lee2019SpherePHDAC}&57K&59.20\\
    &Spherical GCN~\cite{Yang2020RotationEG}&60k&60.72\\
    &PanoSwin-Tiny~\cite{ling2023panoswin}&30M&75.01\\
    &SPHTR~\cite{cho2024sampling}&60k&58.21\\
    \bottomrule
    \end{tabular}}
    \vspace{-5pt}
\end{table}

\vspace{-15pt}
\subsection{Euclidean space}
\label{sec3.1}
\vspace{-5pt}

The most popular planar projection, ERP, supports the networks designed for perspective images. However, the equal spacing of longitudes and latitudes in ERP introduces severe distortions, especially at the two poles. These distortions lead to a degradation in performance when 2D-designed networks are directly applied to ERP images. Therefore, there exist numerous specific representation learning methods for ERP images. 

\begin{figure*}[t!]
  \centering
  \includegraphics[width=0.90\linewidth]{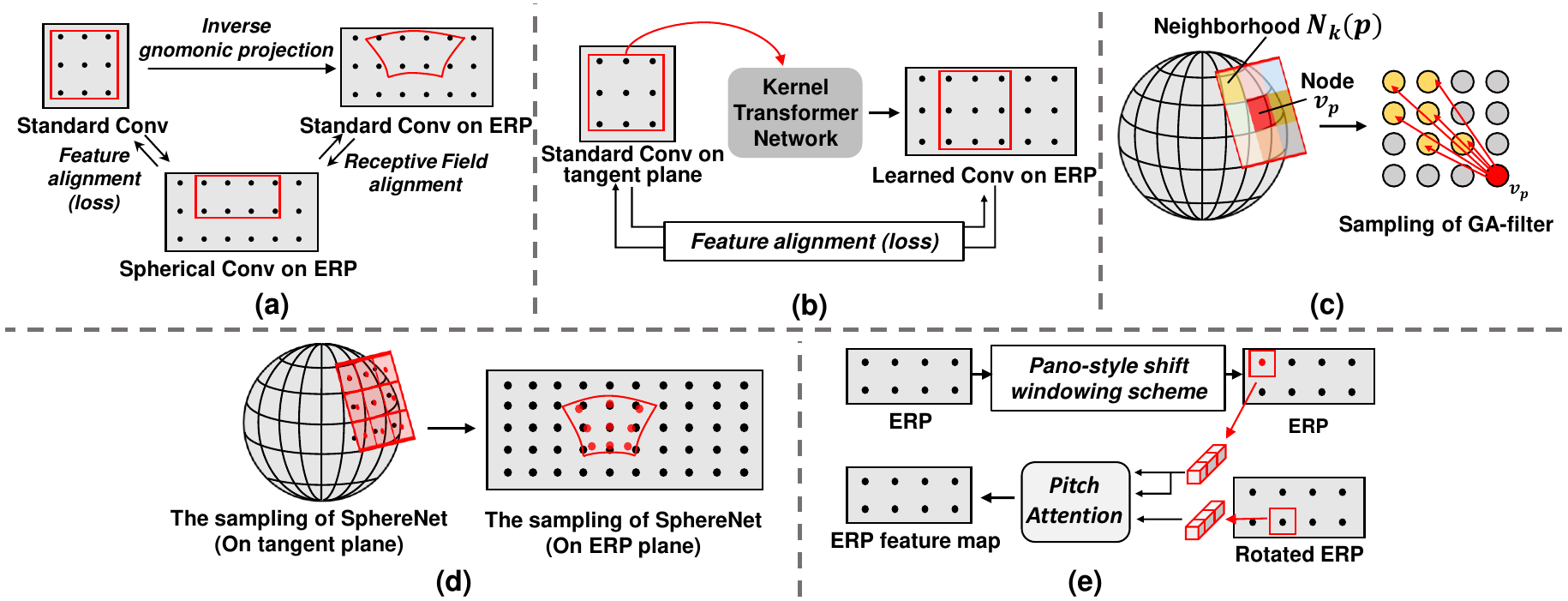}
      \vspace{-8pt}
  \caption{Illustrations of representative representations on Euclidean space: (a) Spherical convolution~\cite{Su2017LearningSC}; (b) Kernel transformer network~\cite{Su2019KernelTN}; (c) Geometry-aware convolution~\cite{Khasanova2019GeometryAC}; (d) SphereNet~\cite{coors2018spherenet}; (e) PanoSwin~\cite{ling2023panoswin}.} 
    \label{fig:odi_representation_eu}
    \vspace{-8pt}
\end{figure*}

Spherical Convolution~\cite{Su2017LearningSC}, as shown in Fig.~\ref{fig:odi_representation_eu}(a), which proposes to leverage the knowledge of networks trained on perspective images to guide the ERP representation learning. Specifically, Spherical Convolution dynamically modifies the shapes of convolutional kernels, promoting alignment between the learned features from ERP data and the perspective features extracted from the associated tangent plane. Furthermore, besides the constraints of feature alignment, KTN~\cite{Su2019KernelTN},~\cite{Su2021LearningSC} introduces a trainable network to adapt source CNNs trained on perspective images to ERP images, focusing on the transformation of convolution kernels, as illustrated in Fig.~\ref{fig:odi_representation_eu}(b). GA-CNN~\cite{Khasanova2019GeometryAC} (Fig.~\ref{fig:odi_representation_eu}(c))  proposes the graph-based convolution filters, whose size and shape adaptively depend on the position in ERP image. Meanwhile, SphereNet~\cite{coors2018spherenet} (Fig.~\ref{fig:odi_representation_eu}(d)) wraps the convolution filters to convert the sampling locations on ERP images into tangent plane around the sphere, effectively improving the distortion invariance. Recently, inspired by the success of vision transformers~\cite{dosovitskiy2020image},~\cite{Liu2021SwinTH}, PanoSwin~\cite{ling2023panoswin} proposes a 
transformer-based representation learning method for ERP images, as depicted in Fig.~\ref{fig:odi_representation_eu}(e). Based on Swin Transformer~\cite{Liu2021SwinTH}, PanoSwin introduces the pano-style shift windowing scheme and pitch attention to address the edge discontinuity and distortions of ERP images.

\begin{figure}[!t]
  \centering
  \includegraphics[width=0.85\linewidth]{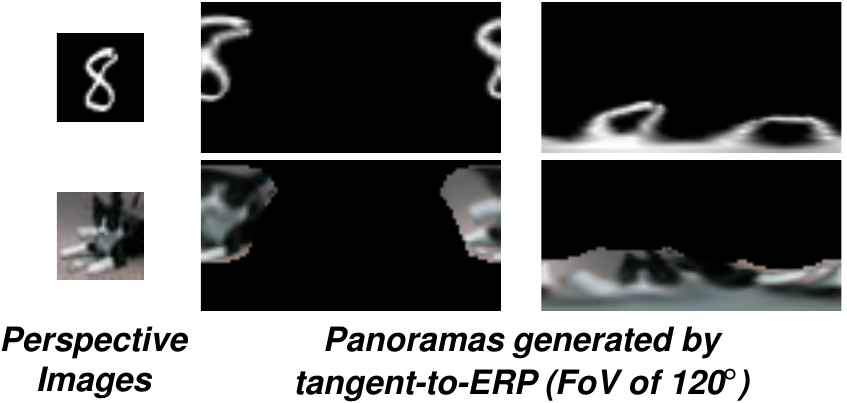}
  \vspace{-5pt}
\caption{SPH-MNIST (top) \& SPH-CIFAR10 (down).} 
    \label{fig:spherical_images_for_classification}
    \vspace{-15pt}
\end{figure}

\begin{figure*}[!t]
  \centering
  \includegraphics[width=0.90\linewidth]{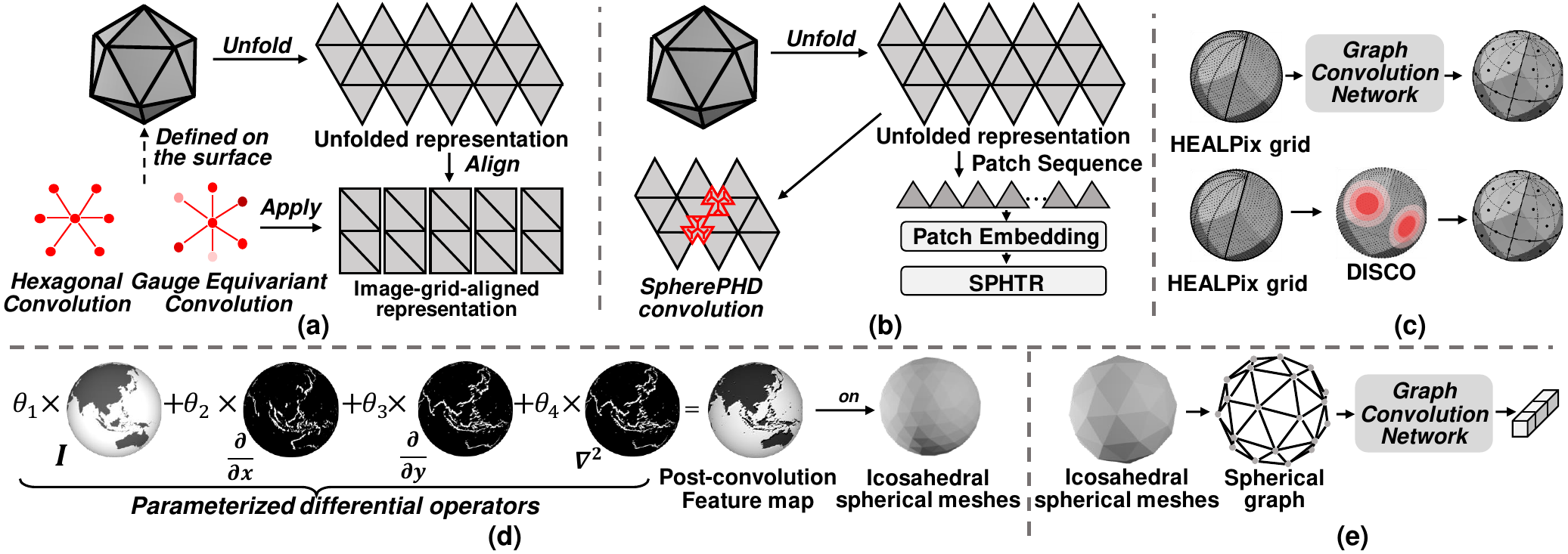}
      \vspace{-8pt}
  \caption{Illustrations of representative ODI representation learning methods on non-Euclidean space: (a) HexNet~\cite{Zhang2023HexNetAO} and Gauge CNN~\cite{cohen2019gauge}; (b) SpherePHD~\cite{Lee2019SpherePHDAC} and SPHTR~\cite{cho2024sampling}; (c) DeepSphere~\cite{Defferrard2020DeepSphere} and DISCO~\cite{OcampoPM23}; (d) Parameterized differential operators~\cite{JiangHKPMN19,Shen2021PDOeS2CNNsPD}; (e) Spherical GCN~\cite{Yang2020RotationEG}.} 
    \label{fig:odi_representation_noneu}
    \vspace{-10pt}
\end{figure*}

\vspace{-15pt}
\subsection{Non-Euclidean space}
\vspace{-5pt}
\label{sec3.2}

As the raw spherical data possesses a non-Euclidean spherical data structure, some methods have explored the Non-Euclidean convolution filters in the spherical domain. They represent spherical images as icosahedron meshes~\cite{Liu2018DeepL3} or Hierarchical Equal Area iso-Latitude Pixelation (HEALPix) grids~\cite{Gorski2005HEALPixAF} for processing. However, standard convolution filters are not applicable to Non-Euclidean representations. Therefore,~\cite{cohen2019gauge},~\cite{Zhang2023HexNetAO} (Fig.~\ref{fig:odi_representation_noneu}(a)) propose similar solutions that are to unfold icosahedron meshes and align its components to the standard image grids. Similarly, SpherePHD~\cite{Lee2019SpherePHDAC} and SPHTR~\cite{cho2024sampling} also employed the unfolded icosahedron meshes as the input of the proposed CNN and Transformer, respectively. However, as illustrated in Fig.~\ref{fig:odi_representation_noneu}(b), the CNN implementation of SpherePHD,~\ie, convolution and pooling layers, and the patch embedding of SPHTR, are applied onto the adjacent triangle. By contrast,~\cite{JiangHKPMN19},~\cite{Yang2020RotationEG},~\cite{shakerinava2021equivariant},~\cite{Shen2021PDOeS2CNNsPD} directly employ the icosahedron meshes as the input without unfolding. Especially,~\cite{JiangHKPMN19},~\cite{Shen2021PDOeS2CNNsPD} parameterize convolution kernels as a linear combination of differential operators for processing the meshes (Fig.~\ref{fig:odi_representation_noneu}(d)), while~\cite{Yang2020RotationEG} converts the icosahedron mesh into a spherical graph and designs a rotation-equivariant graph convolution network (GCN) to extract features (Fig.~\ref{fig:odi_representation_noneu}(e)). 
In addition,~\cite{eder2020tangent} verifies that it is effective to extract features from less-distorted tangent planes of a subdivided icosahedron through the standard CNNs. For learning ODI representations from the HEALPix gird input, as shown in Fig.~\ref{fig:odi_representation_noneu}(c), DeepSphere~\cite{Defferrard2020DeepSphere} constructs a rotation equivariant graph convolution network while DISCO~\cite{OcampoPM23} proposes a novel equivariant and scalable group convolution based on the spherical coordinates.

\vspace{-13pt}
\subsection{Discussion and Potential}
\vspace{-5pt}
\label{sec3.3}
In Tab.~\ref{tab:representation_comparison}, we show the benchmark classification results of representative spherical representation learning methods. Since there are currently no dedicated classification datasets designed for spherical image classification, existing methods treat perspective image datasets, such as MNIST~\cite{deng2012mnist} and CIFAR10~\cite{Krizhevsky2009LearningML}, as tangent patches, and then convert them into panoramas through inverse gnomonic projection. Data augmentation is then performed through rotation, as depicted in Fig.~\ref{fig:spherical_images_for_classification}. To the best of our knowledge, the tiny version of PanoSwin~\cite{ling2023panoswin} has attained the best results up to now.

From the aforementioned analysis, most existing Euclidean space-based methods~\cite{Su2017LearningSC},~\cite{coors2018spherenet},~\cite{Su2019KernelTN},~\cite{Su2021LearningSC} have explored the potential of connecting ERP and perspective images. They propose distortion-aware convolutions based on the geometric relationship in the inverse sphere-to-plane projection to alleviate the issue of 2D off-the-shelf network failure caused by ERP distortion. However, these distortion-aware convolution filters bring substantial computational costs and suffer from the local receptive field. Inspired by the success ofPanoSwin~\cite{ling2023panoswin}, the desired direction for Euclidean space-based methods is to specifically design distortion-aware attention mechanisms. For the non-Euclidean space-based methods, existing methods primarily focus on hand-crafted CNNs for unfolded icosahedron meshes or applying rotation equivariant graph convolution networks on the graphs constructed from icosahedron meshes. As transformer-based graph models have shown their power~\cite{Chen2022AUA},~\cite{Chen2022StructureAwareTF},~\cite{Zhu2023HierarchicalTF}, it could be beneficial to infer these transformer-based graph models and customize spherical graph Transformers for non-Euclidean ODI representation learning.

\vspace{-10pt}
\section{Optimization Strategies}
\vspace{-8pt}
\label{sec4}

In this section, we will introduce essential optimization strategies applied to omnidirectional vision beyond the supervised learning strategy. To demonstrate the evolution of optimization strategies, we will conduct analyses across multiple tasks.

\begin{figure}[!t]
  \centering
      \includegraphics[width=0.9\linewidth]{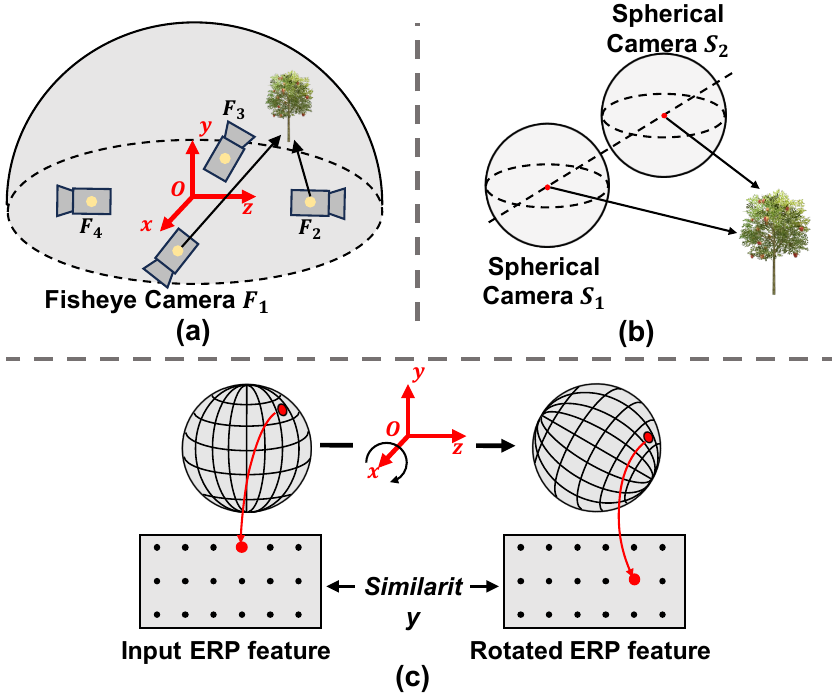}
      \vspace{-10pt}
  \caption{The input in unsupervised or semi-supervised optimization strategy: (a) Four wide-FoV fisheye images; (b) Two spherical images captured with the horizontal baseline; (c) One spherical image and rotated one.} 
    \label{fig:unsupervised}
    \vspace{-15pt}
\end{figure}

\vspace{-10pt}
\subsection{Unsupervised/Semi-supervised Learning}
\vspace{-5pt}
\label{sec4.1}

Data scarcity problem occurs on ODIs due to the insufficient yet costly panorama annotations. This problem is commonly addressed by semi-supervised learning or unsupervised learning that can take advantage of abundant unlabeled data to enhance the generalization capacity. In existing unsupervised\footnote{Also called self-supervised.} and semi-supervised learning methods for ODI data, as shown in Fig.~\ref{fig:unsupervised}, there can be divided into three main categories according to the input data: 1) The input is multiple wide field-of-view (wide-FoV) fisheye images (Fig.~\ref{fig:unsupervised}(a)); 2) The input is a pair of panoramas captured from different view positions (Fig.~\ref{fig:unsupervised}(b)); 3) The input is a panorama and its rotated one (Fig.~\ref{fig:unsupervised}(c)). For the first category, representative works are~\cite{Lee2022SemiSupervised3D},~\cite{Chen2023UnsupervisedOE}.~\cite{Lee2022SemiSupervised3D} calculates the unsupervised photometric loss according to the overlapping areas of any two fisheye cameras. In U-OmniMVS~\cite{Chen2023UnsupervisedOE}, four wide-FoV fisheye images with the FoV of 220$^\circ$ are split into two back-to-back pairs and projected into two ERP images. After that, the pseudo-stereo supervision is established on two ERP images following the spherical pseudo-stereo matching. The pipeline of second category is to synthesize novel views from a single input RGB image combined with the estimated depth map or depth ground truth, and then employ the corresponding areas between input view and generated view, calculated by spherical geometry constraint, as the supervision~\cite{Zioulis2019SphericalVS},~\cite{Liu2021PanoSfMLearnerSM},~\cite{Tran2021SSLayout360SI},~\cite{Kulkarni2022360FusionNeRFPN},~\cite{Hsu2021MovingIA},~\cite{Hara2022EnhancementON}. The third category,~\cite{Bhandari2022LearningOF},~\cite{Abdelaziz2021Rethinking3I} aims to maximize the correlations between the representations from original input panoramas and corresponding rotated panoramas.

\begin{figure}[!t]
  \centering
        \includegraphics[width=0.85\linewidth]{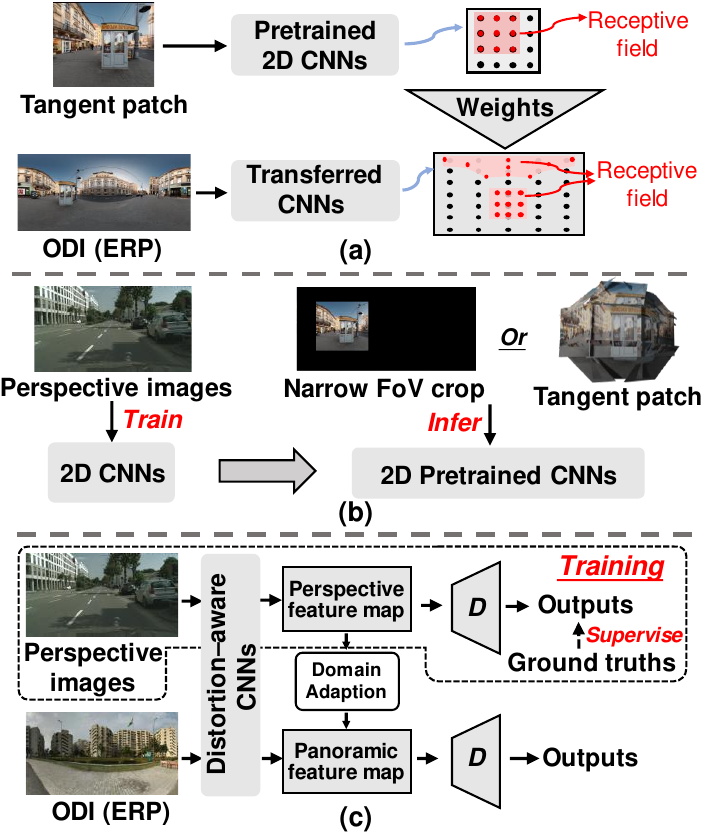}
        \vspace{-8pt}
  \caption{The illustrations of three directions of transfer learning optimization strategy: (a) kernel-level, (b) data-level (with TP as the example); (c) feature-level.} 
    \label{fig:optimization_transfer}
    \vspace{-12pt}
\end{figure}

\vspace{-15pt}
\subsection{Transfer Learning}
\vspace{-5pt}
\label{sec4.2}

Transfer learning strategy can leverage the wealthy resource of perspective images to facilitate rapid progress in the field of omnidirectional vision. Meanwhile, transfer learning enables the adaptation of knowledge from perspective image domain to the ODI domain, thereby mitigating the challenges posed by limited annotated data and high data acquisition costs in omnidirectional vision. Transfer learning on ODIs can be divided into three major directions: kernel-level transfer, data-level transfer, and feature-level transfer. For kernel-level transfer, the representative works~\cite{coors2018spherenet},~\cite{Tateno2018DistortionAwareCF},~\cite{Su2019KernelTN},~\cite{bhandari2021revisiting},~\cite{Su2021LearningSC},~\cite{artizzu2021omniflownet} follow the idea of training a standard CNN on the abundant perspective images, and then performing a sampling grid transformation on the convolution kernel based on the geometric relationship between ERP and perspective projection (Fig.~\ref{fig:optimization_transfer}(a)). Using kernel-level transfer,
it is possible to use the weights of pre-trained standard CNNs and make the networks distortion-aware. Data-level transfer methods, as visualized in Fig.~\ref{fig:optimization_transfer}(b), first convert distorted ERP images into narrow FoV slices~\cite{yu2023panelnet},~\cite{yang2020omnisupervised} or less-distorted patches~\cite{zheng2023both},~\cite{eder2020tangent},~\cite{zhang2022360}. Subsequently, networks pre-trained on perspective images are directly employed on slices or patches. Compared with the above two directions, feature-level transfer learning strategy is more popular and mainly seen in unsupervised domain adaptation methods for panoramic semantic segmentation~\cite{Yang2021CapturingOC},~\cite{Zhang2021TransferBT},~\cite{ma2021densepass},~\cite{Zhang2022BendingRD},~\cite{Zheng2023LookAT}. Following Fig.~\ref{fig:optimization_transfer}(c), they adapt extracted semantic features in the labeled pinhole image domain to the unlabeled panoramic image domain with several shared classes.

\begin{figure}[t!]
\centering
\includegraphics[width=0.85\linewidth]{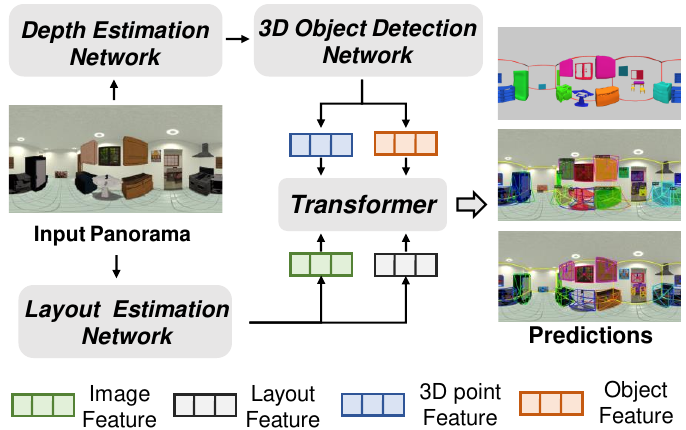}
\vspace{-5pt}
\caption{The representative multi-task learning approach: PanoContext-Former~\cite{Dong2023PanoContextFormerPT}.}
    \label{fig:multitask_learning}
    \vspace{-15pt}
\end{figure}

\vspace{-15pt}
\subsection{Multi-task Learning}
\label{sec4.3}
Multi-task learning strategy emphasizes sharing representations between the related tasks can increase the generalization capacity of the models and improve the performance on all involved tasks. V-CNN~\cite{Xu2020ViewportBasedCA} handles four tasks: camera motion detection, viewport proposal, saliency prediction, and visual quality assessment score rating to investigate the correlations between videos' visual quality and the head movement and eye movement of subjects on watching omnidirectional videos. Similar to V-CNN,~\cite{Zhou2022OmnidirectionalIQ} proposes an interesting direction that employs an auxiliary distortion discrimination task to reduce the impact of distortions on the quality assessment task learning. This provides a good inspiration for mitigating the distortion issues. Pano-SfMLearner~\cite{Liu2021PanoSfMLearnerSM} shares the information between three related tasks,~\ie, depth estimation, camera motion estimation, and semantic segmentation, to mutually promote each other. Similarly, for scene understanding based on ODIs, there exist several methods~\cite{Zhang2021DeepPanoContextP3,Dong2023PanoContextFormerPT} that improve the performance on each sub-task by sufficient exploitation of contextual information and geometric relationships between scene objects. In particular, we provide an illustrative diagram of~\cite{Dong2023PanoContextFormerPT} (Fig.~\ref{fig:multitask_learning}) to better understand the advantages of multi-task learning. It can be seen that integrating feature representations obtained from different tasks can utilize the rich context information and better understand the components of the scene.

\vspace{-15pt}
\subsection{Deep Reinforcement Learning}
\begin{figure}[t!]
\centering
\includegraphics[width=0.90\linewidth]{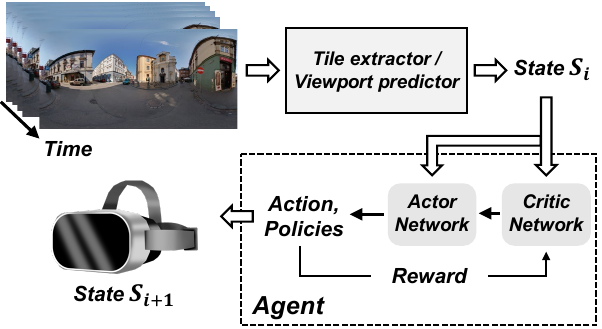}
    \vspace{-3pt}
\caption{The common framework of DRL-based rate adaptation for 360$^\circ$ video streaming.}
    \label{fig:reinforcement_learning}
    \vspace{-10pt}
\end{figure}

\vspace{-5pt}
\label{sec4.4}
Reinforcement learning (RL) aims to computationally determine the states of a specific system and execute an action to optimize an accumulated reward for the system. Deep reinforcement learning (DRL) strategy leverages the deep learning approaches to enhance reinforcement learning's ability to tackle decision-making challenges previously deemed unsolvable, particularly those involving high-dimensional state and action spaces. As omnidirectional videos (ODVs) are always with high resolution, limited network bandwidth is challenging for transmitting ODVs and proving high Quality of Experience (QoE) of users. Thus, it is to valuable to explore how to effectively allocate the bit-rates for different content of ODVs using DRL strategy. To better understand how it works, we present a common framework in Fig.~\ref{fig:reinforcement_learning}. As a natural sequential decision-making problem, bit-rate allocation can be naturally optimized with the DRL strategy. Specially, due to the limited FoV of human eyes, an individual can only see a single viewport of each ODV frame at any given time. Therefore, existing DRL-based ODV streaming and caching methods~\cite{Ban2020MA360MD,Zhang2019DRL3603V,Park2021AdaptiveSO,Jiang2020ReinforcementLB,Kan2021RAPT360RL,Maniotis2020ViewportAwareDR,Li2022SAD360SV} adaptively allocate bit-rates based on the viewports. Specifically, they first predicting a user's viewport, then assigning tile granularities according to this viewport, and subsequently allocate higher or lower bit-rates to different tiles. Ultimately, through DRL strategies based on QoE metrics, the reward of executing actions is calculated, facilitating the adjustment of allocation.
Moreover, for saliency prediction,~\cite{MaiXu2021SaliencyPO} predicts head fixation through DRL strategy by interpreting the trajectories of head movements as discrete actions. Meanwhile,~\cite{Deng2021LAUNetLA} employs the DRL strategy to select up-scaling factors of super-resolution adaptively based on unevenly distributed pixel density in the ERP images.
\begin{figure*}[!t]
  \centering
  \includegraphics[width=1\linewidth]{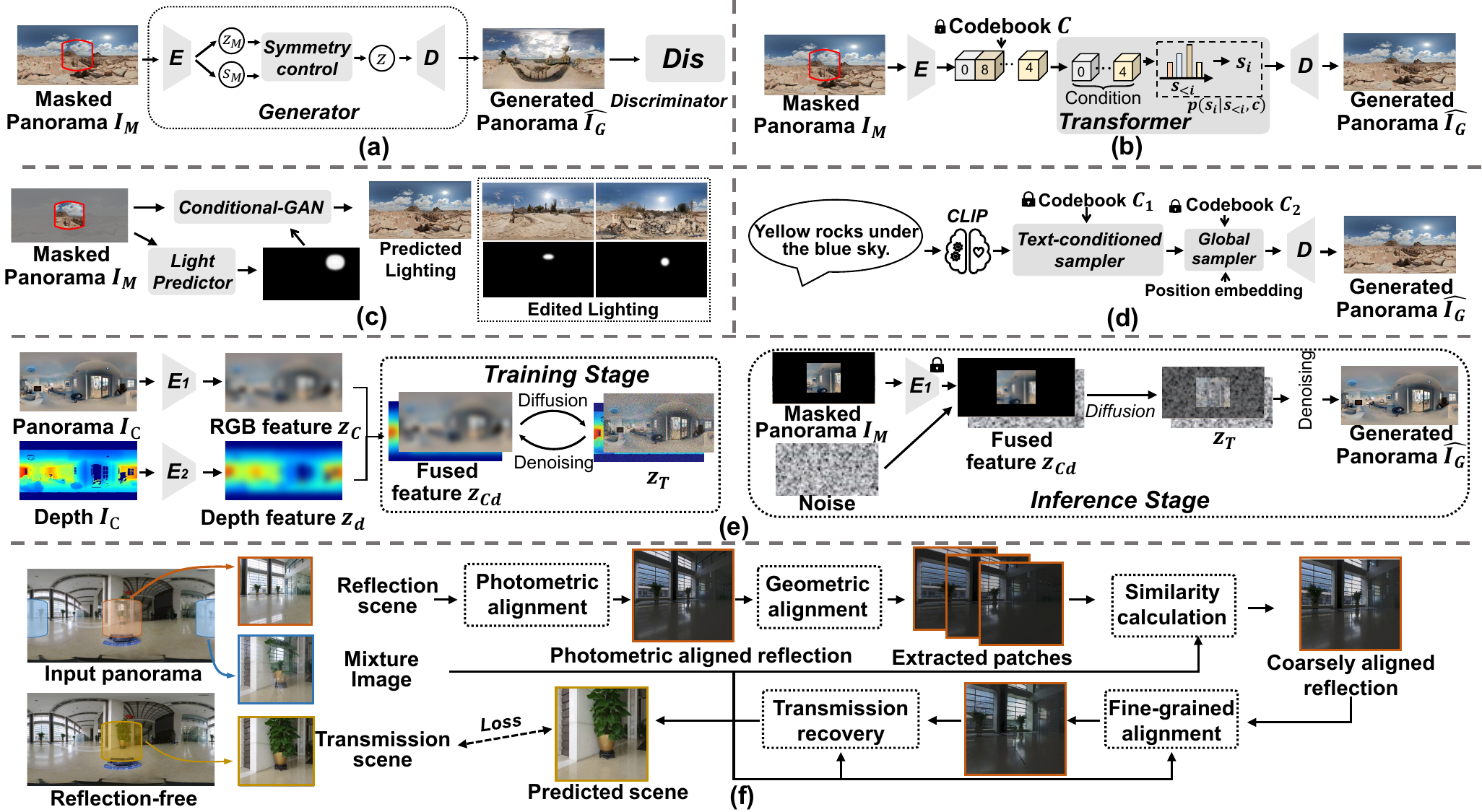}
  \vspace{-8pt}
  \caption{The illustrations of representative networks for ODI Generation: (a) GAN-based outpainting~\cite{hara2020spherical}; (b) Transformer-based outpainting~\cite{Akimoto2022DiverseP3}; (c) GAN-based lighting estimation~\cite{Dastjerdi2023EverLightIE}); (d) Transformer-based text-guided generation~\cite{Chen2022Text2Light}; (e) Diffusion-based outpainting~\cite{wu2024panodiffusion}; (f) Matching-based reflection removal~\cite{hong2021panoramic}.} Not all details of the architecture are depicted.
    \label{fig:generation}
    \vspace{-12pt}
\end{figure*}

\vspace{-10pt}
\section{Omnidirectional Vision Tasks}
\label{sec5}
In this work, we primarily focus on three key aspects of prevailing DL techniques for omnidirectional vision: visual enhancement (Sec.~\ref{sec5.1}), scene understanding (Sec.~\ref{sec5.2}), and 3D geometry and motion estimation (Sec.~\ref{sec5.3}). For each aspect, we select some of the most influential research fields, as introduced in Fig.~\ref{fig:survey}. We aim to facilitate a quick understanding of the development for newcomers through a summary and analysis of representative methods mostly published in top-tier conferences or journals.

\vspace{-10pt}
\subsection{Visual Enhancement}
\label{sec5.1}

\subsubsection{ODI Generation}
\label{sec5.1.1}

\noindent\textbf{ODI inpainting/outpainting:} Image inpainting and outpainting are conceptually opposite, where inpainting aims to fill missing or damaged parts of an image while outpainting requires extending an image beyond its original boundaries. Especially, current efforts on ODI generation are primarily focused on outpainting, which extrapolates a complete high-quality panorama from the given narrow FoV (NFoV) images (typically masked panoramas). Notably, compared to perspective images, ODI outpainting faces the challenge, that is the continuity at the left and right sides of panoramas. 

The pioneering work, 360IC~\cite{akimoto2019360} introduces an insightful approach. By harnessing the cylinder projection property of ERP and exploiting continuity on both ends, 360IC shifts the given NFoV content from the center towards the edges and employs two-stage conditional GANs~\cite{Mirza2014ConditionalGA} to generate the missing area. This innovative technique effectively transforms the outpainting challenge into an easier inpainting task. Furthermore, as only given content on both sides is continuous in 360IC, SIG-SS~\cite{hara2020spherical} proposes circular shift and circular padding to control kinds of scene symmetries and maintain continuity on whole regions of both sides. Especially, circular padding strategy is popular in panorama outpainting methods to ensure the wraparound consistency of generated panoramas~\cite{Dastjerdi2022GuidedCG},~\cite{Liao2022CylinPaintingS3},~\cite{Wang2023360DegreePG},~\cite{Shum2023Conditional3I},~\cite{Chen2022Text2Light},~\cite{wang2024customizing}. For instance, ~\cite{wu2024panodiffusion} applies circular padding as data augmentation. Moreover, circular inference~\cite{Akimoto2022DiverseP3},~\cite{Ai2024Dream360DA} are common in transformer-based methods. Besides, Cylin-Painting~\cite{Liao2022CylinPaintingS3} constructs a GAN-based framework with the cylinder-style convolution layers and exploits collaborations between inpainting and outpainting, which are geometrically related within a seamless cylinder.

Inspired by recent advancements in generative models based on perspective images, numerous panorama outpainting works also strive to generate more diverse plausible panoramas based on given NFoV images, as shown in Fig.~\ref{fig:generation}. Based on conditional VAE~\cite{Sohn2015LearningSO}, SIG-SS~\cite{hara2020spherical} utilizes the scene symmetry intensity of global structure to control the generated content (Fig.~\ref{fig:generation}(a)). Omnidreamer~\cite{Akimoto2022DiverseP3} and Dream360~\cite{Ai2024Dream360DA} (Fig.~\ref{fig:generation}(b)), based on Taming transformer~\cite{Esser2020TamingTF}, generate high-resolution, diverse and semantically consistent results with auto-regressive transformers. To improve the generation quality and control the generation styles, ImmerseGAN~\cite{Dastjerdi2022GuidedCG} introduces a class-driven discriminative network into CoModGAN~\cite{zhao2021large} for editing the scene classes. Meanwhile, built on the latest latent diffusion model~\cite{Rombach2021HighResolutionIS}, PanoDiffusion~\cite{wu2024panodiffusion} takes panoramic depths and NFoV images as the joint inputs and presents the rotation-augmented diffusion process to generate well-structured results (Fig.~\ref{fig:generation}(e)).

For the inpainting task, whether based on regular masks or free-form masks, existing methods~\cite{han2020piinet},~\cite{shang2022viewport},~\cite{Yu2024PanoramicII} focus on NFoV viewports of ODIs.~\cite{han2020piinet},~\cite{Yu2024PanoramicII} convert a single panorama into six CP patches and directly apply the perspective inpainting network on these less-distorted CP patches. Differently,~\cite{shang2022viewport} perform generative adversarial inpainting on both panorama and the set of less-distorted NFoV viewports and fuse the features from ERP and viewports to improve the generation performance.

\noindent\textbf{Text-Guided Generation:} Text guidance can provide a more interactive experience, which is crucial for practical applications such as AR/VR. Text2Light~\cite{Chen2022Text2Light} utilizes the pre-trained Contrastive Language-Image Pre-training (CLIP) model~\cite{Radford2021LearningTV} and auto-regressive transformer-based generative models~\cite{Esser2020TamingTF} to learn the text-conditioned samplers for modeling the ODI data distributions and synthesizing panoramas patch-by-patch (Fig.~\ref{fig:generation}(d)). Similarly, PanoGen~\cite{li2023panogen} first converts an ODI into multiple tangent patches and uses the vision-language model BLIP-2~\cite{li2023blip} to generate text captions for each patch. Subsequently, these annotations, along with a SoTA text-to-image diffusion model~\cite{Rombach2021HighResolutionIS}, are recursively used to generate new tangent patches while ensuring semantic consistency across patches. Finally, the generated tangent patches are integrated through stitching to produce the final ERP-format ODI.~\cite{wang2024customizing} fine-tunes MultiDiffusion~\cite{BarTal2023MultiDiffusionFD} based on the Low-Rank Adaptation (LoRA)~\cite{hu2022lora} for customizing panoramas according to the text. Differently, built on the latent diffusion model, AOG-Net~\cite{Lu2024AutoregressiveOO} and PanoDiff~\cite{Wang2023360DegreePG} focus on the text-guided panorama outpainting task, which generate the complete panoramas with the guidance of NFoV images and text prompts. Notably, both AOG-Net and PanoDiff support input from multiple NFoV images. Particularly, the NFoV images in AOG-Net are registered, while in PanoDiff the NFoV images can be unregistered. This owes to PanoDiff introduces an angle prediction branch to predict the shooting positions of multiple NFoV images. 

\noindent\textbf{Lighting Estimation:} It is to estimate the high dynamic range (HDR) illumination of the surrounding environment and approximate realistic lighting effects while inserting virtual objects. Specifically, the most popular lighting representation is environment maps, which are ERP format ODIs with promising realistic reflections. For lighting estimation with the environment maps, there exist numerous learning-based works~\cite{Gardner2017LearningTP,hold2019deep,Song2019NeuralIL,Somanath2020HDREM,Wang2022StyleLightHP,Dastjerdi2023EverLightIE} to directly generate the HDR panoramas from the given low dynamic range (LDR), limited field-of-view (LFoV) images. As the pioneer work, Gardner~\etal~\cite{Gardner2017LearningTP} presented an end-to-end training approach with a two-stage network to predict the HDR environment maps. In the first stage, a deep convolution network is trained to anticipate the spatial arrangement of light sources within a complete scene from a given LDR LFoV input. By fine-tuning the model on the HDR panorama dataset, this network can predict the HDR environment map with the light intensities from the given LDR LFoV image. Differently from the fine-tuning strategy for mitigating task difficulty in~\cite{Gardner2017LearningTP},~\cite{hold2019deep} proposes a step-by-step sky modeling approach. This method sequentially learns the latent code for HDR sky panorama reconstruction, trains the network for LDR-to-HDR sky panorama conversion, and predicts HDR sky panorama from LDR LFoV images using the previous latent codes and conversion network. In contrast, Neural Illumination~\cite{Song2019NeuralIL} provides a universal procedure for the lighting estimation with three steps: 1) converting LDR LFoV inputs into partial LDR panoramas, 2) completing partial LDR panoramas, and 3) performing LDR-to-HDR estimation. With the strong GANs for scene completion~\cite{goodfellow2014generative}, EnvMapNet~\cite{Somanath2020HDREM} directly synthesizes complete HDR panorama from the partial LDR panoramas. Recently, editable lighting estimation with an HDR environment map has been attractive, allowing for flexible control over light intensity. For instance, StyleLight~\cite{Wang2022StyleLightHP} utilizes dual-StyleGAN~\cite{Yang2022PasticheME} to predict HDR environment maps from partial LDR panoramas and achieves editable lighting through GAN inversion. To avoid expensive GAN inversion techniques, EverLight~\cite{Dastjerdi2023EverLightIE} generates HDR environment maps via combining GAN and parametric lighting models, and controls the illumination based on predicted lighting parameters (See Fig.~\ref{fig:generation}(c)).

Besides the HDR environment map-based lighting estimation works, other lighting representations have also been well explored,~\eg, parametric light sources~\cite{HoldGeoffroy2016DeepOI},~\cite{Gardner2019DeepPI},~\cite{Zhang2019AllWeatherDO}, spatially-varying spherical harmonics~\cite{Garon2019FastSI},~\cite{Zhu2021SpatiallyVaryingOL}, and spherical gaussians~\cite{Li2020InverseRF},~\cite{Zhan2020EMLightLE},~\cite{Zhan2021GMLightLE}, etc. These methods aim to employ deep networks to predict a compact set of lighting parameters (~\eg, sun position, lighting intensities, sun shape, sky turbidity) or spherical harmonics coefficients, which are used to reconstruct HDR illumination maps with the ERP format. From the aforementioned analysis, it is evident that many methods related to lighting estimation share overlaps with panorama outpainting. However, pure panorama outpainting methods primarily focus on generating complete LDR panoramas from masked LDR panorama images, emphasizing diversity and realism in the generated regions. By contrast, lighting estimation concerns itself with producing HDR results. In addition to plausible textures, it also places emphasis on shading and strong cast shadow.

\noindent \textbf{Reflection removal:} It aims to analyze reflection content and recover the transmission scene from reflection-contaminated images (commonly referred to as mixed images). As ODIs capture the surrounding environment in a single shot, they inherently provide the reflection scene and mixed scene simultaneously, which can solve the challenge of \textit{content ambiguity}. Recent studies~\cite{hong2021panoramic},~\cite{hong20232},~\cite{han2022zero},~\cite{park2024fully} have demonstrated the potential of ODIs for reflection removal. The pioneering work~\cite{hong2021panoramic} introduced a representative pipeline, as shown in Fig.~\ref{fig:generation}(f), which consists of identifying the mixed image (an NFOV region of ODI) based on user interactions, locating 
corresponding glass reflection image, and learning to recover the transmission scene through matching cropped reflection and mixed images. Recently, Park~\etal~\cite{park2024fully} utilized the similarities among attention windows in transformer blocks~\cite{dosovitskiy2020image} to ensure the locations of reflection and mixed regions and achieved fully-automatic end-to-end learning. Additionally, ZS360~\cite{han2022zero} employs zero-shot learning, assuming the ODI center aligns with the reflection glass center, to recover the transmission scene without requiring large-scale ground truth labels.
\begin{figure}[!t]
  \centering
    \includegraphics[width=0.85\linewidth]{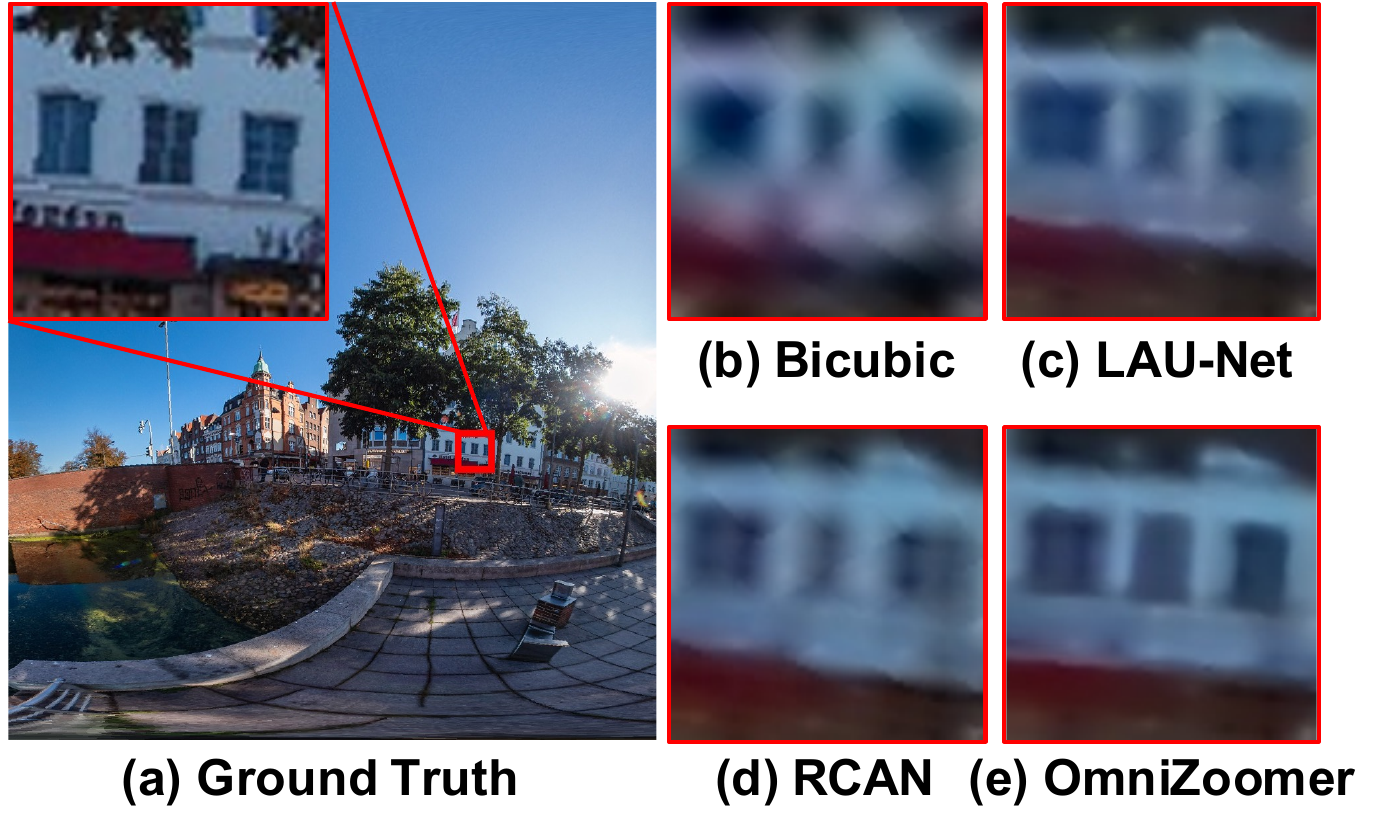}
    \vspace{-10pt}
    \caption{Quantitative results on the SUN360 dataset. (a) High-resolution ground truth ODI. (b) Result of Bicubic interpolation. (3) Result of LAU-Net~\cite{Deng2021LAUNetLA}. (d) Result of RCAN~\cite{zhang2018image}. (e) Result of OmniZoomer~\cite{cao2023omnizoomer}.} 
    \label{fig:SR}
    \vspace{-15pt}
\end{figure}

\noindent\textbf{Discussion:} As mentioned above, it can be observed that the key directions in existing ODI generation works, compared to planar image generation, focus on ensuring the semantic consistency, content richness, and spherical continuity of generated content at high resolution. Since panoramas can offer high interactivity, especially when applied to head-mounted devices, generating high-resolution and realistic panoramas is of significant research value. Additionally, generating well-decorated indoor panoramas from background photographs, as explored in~\cite{Shum2023Conditional3I}, is with also high practical value. Recently, with the emergence of large-scale image-to-image and text-to-image generation models~\cite{Wang2022OFAUA,Zhang2023AddingCC}, how to efficiently and effectively leverage these powerful models designed for perspective images to advance ODI generation is also meaningful. Furthermore, while most reflection removal studies focus on single regions of an ODI, exploring methods for multi-region reflection detection and removal presents a meaningful challenge.
For the ODV generation, it has received limited attention while 360DVD~\cite{Wang2024360DVDCP} leverages diffusion models for text-guided ODV generation.

\begin{figure}[!t]
  \centering
    \includegraphics[width=\linewidth]{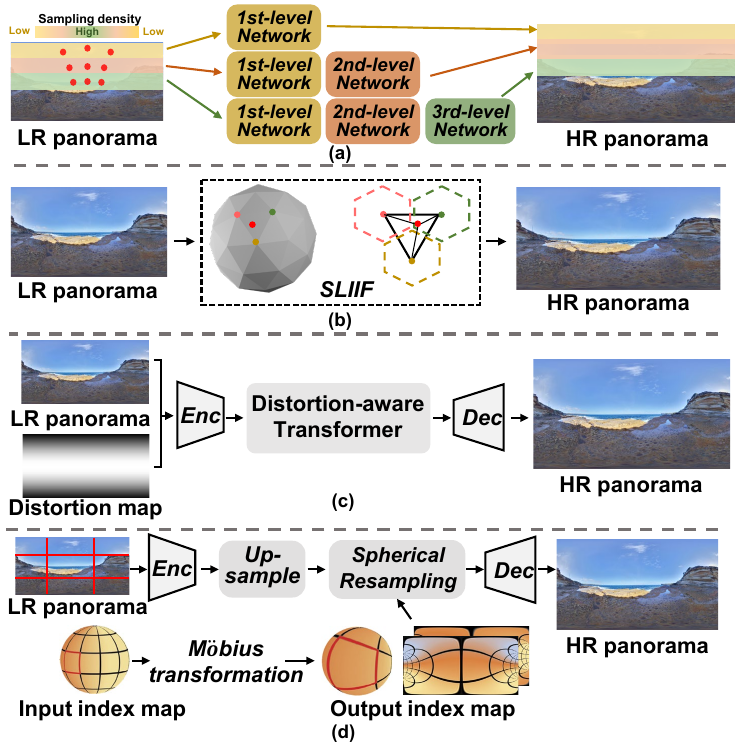}
    \vspace{-20pt}
    \caption{ The illustrations of representative networks for ODI super-resolution: (a)LAU-Net~\cite{Deng2021LAUNetLA}, (b) SphereSR~\cite{yoon2021spheresr}, (c) OSRT~\cite{yu2023osrt}, (d) OmniZoomer~\cite{cao2023omnizoomer}.} 
    \label{fig:SR works}
    \vspace{-10pt}
\end{figure}

\vspace{-5pt}
\subsubsection{Super-Resolution}
\vspace{-3pt}
\label{sec5.1.2}
The Head-Mounted Display (HMD) devices~\cite{rolland2005head} necessitate ODIs with a resolution of at least 21600$\times$10800 for an immersive experience. However, due to the limitation of transmission and storage, such an immersive demand cannot be met by existing camera systems~\cite{ozcinar2019super}. An alternative approach is to capture low-resolution (LR) ODIs and up-sample them into high-resolution (HR) ODIs effectively. Except for designing effective modules for extracting features, the properties of ODIs, such as different distortion levels in different latitudes, spherical projections, and rotation can be utilized to improve the methods' performance and generalization. LAU-Net~\cite{Deng2021LAUNetLA}, pioneering in considering the distortion variance across latitudes for ODI super-resolution (SR), introduces a multi-level network by segmenting an ODI into different latitude bands and upsampling these bands with adaptive factors (See Fig.~\ref{fig:SR works}(a)). Extending beyond SR on the ERP, Yoon~\etal~\cite{yoon2021spheresr} introduced SphereSR to be capable of generating arbitrary projections, \eg, ERP, CP, and perspective views. As illustrated in Fig.~\ref{fig:SR works}(b), the arbitrary projection is accomplished with an implicit function by querying the spherical coordinates. Recently, as shown in Fig.~\ref{fig:SR works}(c), OSRT~\cite{yu2023osrt} designs a transformer architecture with adaptively sampling according to the distortion level, which can mitigate distortion effects in ERP format ODIs. Additionally, OmniZoomer~\cite{cao2023omnizoomer} is designed to super-resolve the ODIs under various transformations, such as zooming in/out and rotation. It utilizes the M\"obius transformation, which is the only bijective transformation with preserved shapes on the sphere (See Fig.~\ref{fig:SR works}(d)). The quantitative results are provided in Fig.~\ref{fig:SR}. Compared with previous methods, OmniZoomer achieves better visual quality, benefiting from its specific design to address ODIs under various transformations. 

% Recently, with the rapid development of diffusion models~\cite{rombach2022high}, OmniSSR~\cite{li2025omnissr} designs a training-free framework that transforms the LR panoramas into a group of tangent patches with low distortions, process them with StableSR in parallel, and finally merge the SR tangent patches with gradient alignment. However, the inference time is about 14 minutes to obtain a 2K SR ODI, which is mainly caused by the multiple denoising steps in Diffusion models and the alignment process.}

\vspace{-5pt}
\subsubsection{Visual Quality Assessment}
\vspace{-3pt}
\label{sec5.1.3}
In today's era of information explosion, billions of images are generated and used every day. Effective visual quality assessment (VQA) can enhance the quality of experience (QoE) of users. Particularly, since ODIs and ODVs can provide interactive experiences with head-mounted displays (HMD) devices, it is crucial to study the quality of ODIs and ODVs to avoid visually unpleasant experiences caused by low-quality content and guide the optimization of broadcasting systems.

Compared to VQA on conventional 2D images and videos, ODI-VQA and ODV-VQA pose greater challenges due to intrinsic factors such as the all-encompassing field-of-view, ultra-high resolution, and geometric deformation resulting from sphere-to-plane projection. Consequently, numerous works have been specifically designed for ODI- and ODV-VQA, tailored to the characteristics of ODIs and ODVs. Notably, VQA can be categorized based on the source of quality scores into subjective and objective methods. Subjective VQA involves collecting reliable human ratings, while objective VQA automatically measures perceptual quality. In this survey, we focus on learning-based methods, thus primarily analyzing objective VQA techniques.
\begin{figure}[!t]
  \centering
   \includegraphics[width=0.95\linewidth]{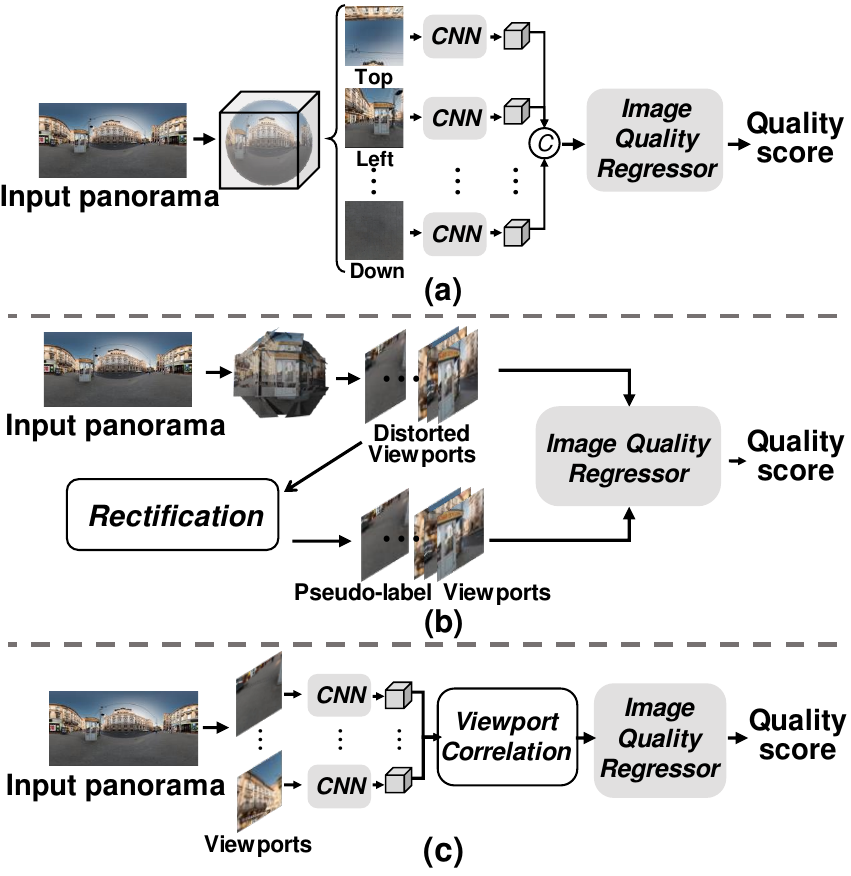}
   \vspace{-10pt}
  \caption{Typical pipelines of NR ODI-VQA: (a) CP patch-based approach; (b) TP patch-based approach; (c) Sequence-based approach.} 
    \label{fig:VQA}
    \vspace{-15pt}
\end{figure}

\noindent\textbf{ODI-VQA.} ODI-VQA can be categorized into full-reference (FR) ODI-VQA and no-reference (NR) ODI-VQA (a.k.a. blind ODI-VQA). The goal of FR ODI-VQA is to measure the difference or fidelity between distorted ODIs and the original ones. However, this task is challenging due to the structural characteristics of ODIs and the complexity of the assessment process. 

Various methods~\cite{Yu2015AFT,Zakharchenko2016QualityMF,Sun2017WeightedtoSphericallyUniformQE} have been developed, leveraging existing 2D image quality metrics such as peak signal-to-noise ratio (PSNR) and structural similarity index (SSIM)\cite{Hor2010ImageQM}. One of the most representative works is S-PSNR\cite{Yu2015AFT}, which extends PSNR to the spherical domain. It samples a limited number of points uniformly distributed on the sphere to calculate the mean error. Additionally, L-PSNR~\cite{Yu2015AFT} assigns larger weights to pixels near the equator. To facilitate quality comparisons across multiple projections, CPP-PSNR~\cite{Zakharchenko2016QualityMF} converts diverse projections into the craster parabolic projection (CPP) for distortion calculation. In contrast, WS-PSNR~\cite{Sun2017WeightedtoSphericallyUniformQE} proposes an efficient and effective weighted-to-spherically uniform method to generate distortion-aware weights at different locations, avoiding complex re-projection operations. In addition, certain methods utilize saliency maps~\cite{Tortorella2017SaliencydrivenOI,Ozcinar2019VisualAO} or head movement maps~\cite{MaiXu2019AssessingVQ} as weight maps to allocate weights in the calculation of PSNR. Similar to the various versions of PSNR, there are also different versions of SSIM for FR ODI-VQA, such as S-SSIM~\cite{SijiaChen2018SphericalSS} and WS-SSIM~\cite{Zhou2018WeightedtoSphericallyUniformSO}. Besides,~\cite{Lim2018VRIN,Kim2020DeepVR} accomplish the quality assessment based on the discriminator between distorted and reference panoramas. The approach involves cropping patches from distorted ERP format ODIs and concurrently estimating weights and quality scores for these patches. The quality score of distorted ODI is then predicted by aggregating the weights and scores of all patches. With network predicted scores and human perception scores, discriminator-based human perception guiders are employed to compare both features of reference and distorted images.

\begin{table}[!t]
    \centering
    \caption{Quantitative comparison of the methods for ODI and ODV VQA task on two popular datasets. The higher the SRCC and PLCC values, the better the performance of the model.
    }
    \vspace{-5pt}
    \label{tab:vqa/comparison-vqa}
    \resizebox{0.9\linewidth}{!}{ 
    \setlength{\tabcolsep}{4pt}
    \begin{tabular}{c|c|c|c|c}
    \toprule
    Datasets&Method  & Type & SRCC $\uparrow$& PLCC$\uparrow$\\
    \midrule
\multirow{6}*{\shortstack{OIQA~\cite{Duan2018PerceptualQA}\\ (ODI)}} 
&WS-PSNR~\cite{Sun2017WeightedtoSphericallyUniformQE}&\multirow{2}*{FR}& 0.3829&0.3678  \\
&WS-SSIM~\cite{Zhou2018WeightedtoSphericallyUniformSO}&& 0.6020&0.3537 \\
\cmidrule{2-5}
&MC360IQA~\cite{Sun2020MC360IQAAM}&\multirow{4}*{NR}&0.9071&0.8925\\
&VGCN~\cite{Xu2020BlindOI}&& 0.9515&0.9584\\
&AHGCN~\cite{Fu2021AdaptiveHC}&& 0.9647&0.9682\\
&Assessor360~\cite{Fu2021AdaptiveHC}&& 0.9802&0.9747\\
\midrule
\multirow{6}*{\shortstack{VQA-ODV~\cite{Xu2020ViewportBasedCA}\\(ODV)}} 
&WS-PSNR~\cite{Sun2017WeightedtoSphericallyUniformQE}&\multirow{2}*{FR}& 0.6232&0.6069  \\
&OV-PSNR~\cite{Gao2022QualityAF}&& 0.7503&0.7483  \\
\cmidrule{2-5}
&MC360IQA~\cite{Sun2020MC360IQAAM}&\multirow{4}*{NR}&0.5106&0.5632\\
&VGCN~\cite{Xu2020BlindOI}&&0.4092&0.4059\\
&BOVQA~\cite{Chai2021BlindQA}&&0.6403&0.5569\\
&CIQNet~\cite{Hu2024OmnidirectionalVQ}&&0.9342&0.9333\\
    \bottomrule
    \end{tabular}}
    \vspace{-16pt}
\end{table}

In many real-world applications, the original ODIs are often unavailable, rendering previous learning-based NR ODI-VQA methods widely applicable. These methods typically follow two main pipelines: \textbf{viewport-based}, and \textbf{sequence-based} approaches, as shown in Fig.~\ref{fig:VQA}. Numerous existing works~\cite{Yang2022TVFormerTV,Zhou2022OmnidirectionalIQ,Sun2020MC360IQAAM,Xu2020BlindOI,Tian2023ViewportSphereBranchNF,Jiang2021CubemapBasedPB,Fu2021AdaptiveHC,Zhou2021NoReferenceQA,wu2023assessor} assess the visual quality of ODIs based on viewports. Viewports represent the undeformed content seen by human eyes during browsing, akin to CP and TP patches. A notable work in this direction is MC360IQA~\cite{Sun2020MC360IQAAM}, which employs CP patches as viewport inputs and presents an image quality regressor that concatenates features of the six viewport images for final quality scores. Building upon MC360IQA, Zhou~\etal~\cite{Zhou2021NoReferenceQA} incorporated an extra distortion discrimination task while using cubemap patches (Fig.~\ref{fig:VQA}(a)) to assess quality scores, facilitating mutual enhancement between the two tasks. In contrast, CPBQA\cite{Jiang2021CubemapBasedPB} learns human attention behavior to weight the quality scores of cubemap patches rather than direct feature concatenation as in~\cite{Sun2020MC360IQAAM,Zhou2021NoReferenceQA}. Furthermore, studies~\cite{Tian2023ViewportSphereBranchNF,Tofighi2023ST360IQNO} extract tangent viewports (Fig.~\ref{fig:VQA}(b)) to predict quality scores. Tian~\etal~\cite{Tian2023ViewportSphereBranchNF} rectified distorted viewports around stitching seams to obtain pseudo-references and calculate quality scores based on differences between distorted and pseudo-reference viewports, while Tofighi~\etal~\cite{Tofighi2023ST360IQNO} sampled tangent viewports from salient parts and directly predict their quality scores.

Recent studies~\cite{Xu2020BlindOI},~\cite{Fu2021AdaptiveHC},~\cite{Yang2022TVFormerTV},~\cite{wu2023assessor},~\cite{Sui2023PerceptualQA} have explored the influence of the observer’s browsing process on ODI-VQA (Fig.~\ref{fig:VQA}(c)). These sequence-based works introduce correlations between different viewports to improve prediction accuracy. One of the first methods to model interactions among different viewports was proposed by Xu~\etal~\cite{Xu2020BlindOI}. They built a spatial viewport graph to model the mutual dependency of viewports and developed a global-local mechanism to learn quality features of global ODI and local viewport sequences simultaneously. However, the limitation of VGCN~\cite{Xu2020BlindOI} and AHGCN~\cite{Fu2021AdaptiveHC} lies in neglecting the dynamic viewport sequence during the browsing process. To address this, TVFormer\cite{Yang2022TVFormerTV} interpolates the head trajectory prediction task to ensure viewport locations and models long-range viewport-to-viewport quality correlation via a transformer-based network. The latest state-of-the-art work, Assessor360~\cite{wu2023assessor}, estimates multiple pseudo-viewport sequences from a given starting point and models viewport-to-viewport correlation for each sequence. The final quality score is the average of quality scores generated for all generated viewport sequences.

\noindent\textbf{ODV-VQA.} In general, assessing ODV quality mirrors ODI quality assessment, especially for FR ODV-VQA. For NR ODV-VQA, Li~\etal~\cite{Li2019ViewportPC} proposed a representative viewport-based CNN approach comprising a viewport proposal network and a viewport quality network. The viewport proposal network generates several potential viewports and their error maps, while the viewport quality network rates the VQA score for each proposed viewport. The final score is calculated as the weighted average of all viewport scores. Another notable work is by Azevedo~\etal~\cite{Azevedo2020AVM}, which considers temporal changes in spatial distortions in ODVs. They fused a set of spatio-temporal objective quality metrics from multiple viewports to learn the final quality score. A similar spatio-temporal consideration strategy is employed in~\cite{Chai2021BlindQA},~\cite{Hu2024OmnidirectionalVQ}. Gao~\etal~\cite{Gao2022QualityAF} modeled the spatial-temporal distortions of ODVs and integrated three existing ODI-QA objective metrics into a novel objective metric, namely OV-PSNR.

\noindent\textbf{Discussion:} Tab.~\ref{tab:vqa/comparison-vqa} shows the state-of-the-art results for ODI-VQA and ODV-VQA on two popular datasets,~\ie, OIQA~\cite{Duan2018PerceptualQA}, and VQA-ODV~\cite{Xu2020ViewportBasedCA}. We report the metrics of Pearson linear correlation coefficient (PLCC) and Spearman
rank-order correlation coefficient (SRCC). To the best
of our knowledge, Assessor360~\cite{wu2023assessor} attains the best results on the OIQA dataset up to now, while CIQNet~\cite{Hu2024OmnidirectionalVQ} outperforms other ODV-VQA methods.

\vspace{-12pt}
\subsection{Scene Understanding}
\vspace{-5pt}
\label{sec5.2}

\subsubsection{Object Detection}
\label{sec5.2.1}
Compared with perspective images, learning-based object detection on ODIs faces two main challenges: 1) traditional convolutional kernels struggle to process the irregular grid structures within ERP; 2) 2D object detection criteria are not well-suited for spherical images.

To tackle the first challenge, some studies have customized network architectures using standard convolution kernels. For example, Tong~\etal~\cite{Tong2019ObjectDF} proposed a multi-scale feature pyramid network, Wang~\etal~\cite{wang2019object} introduced multi-kernel layers, and Shen~\etal~\cite{Shen2021TrainingRP} utilized depth-wise separable convolutions. Additionally, Su~\etal~\cite{Su2017LearningSC} replaced the standard convolution filters of Faster RCNN~\cite{Ren2015FasterRT} with distortion-aware spherical convolutions. Notably, Yang~\etal~\cite{yang2018object} first converted ERP format ODIs into multiple stereographic sub-projections and then employed the 2D object detector to process these less-distorted patches individually.

\begin{figure}[!t]
  \centering
      \includegraphics[width=0.8\linewidth]{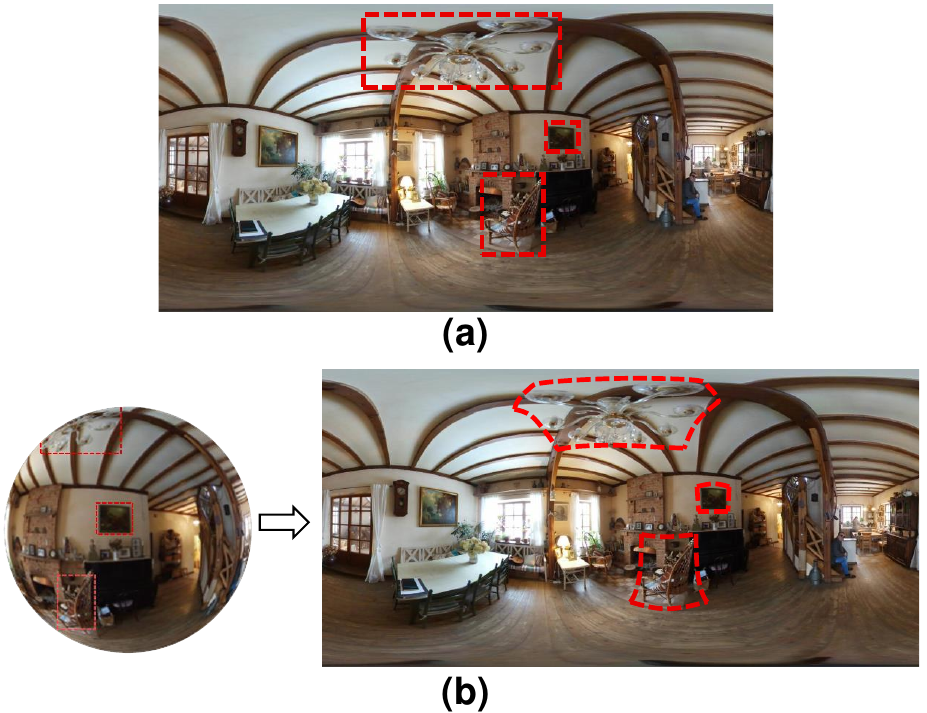}
      \vspace{-10pt}
  \caption{(a) conventional planar bounding boxes (BBs) and (b) field-of-view bounding boxes (FoV-BBs).} 
    \label{fig:objection_detection_BBs}
    \vspace{-15pt}
\end{figure}

For the second challenge, existing methods employ diverse criteria for ODI object detection,~\ie, bounding boxes (BBs) and intersection-over-unions (IoUs). In particular, Coors~\etal~\cite{coors2018spherenet} predicted unbiased bounding boxes (TanBBs) on the tangent plane instead of regressing rectangular BBs on the ERP format ODI. Furthermore, to better align the criteria with the characteristics of spherical imaging in ODI, Spherical Criteria~\cite{zhao2020spherical} proposes spherical BB (Sph-BB) and spherical IoU (Sph-IoU), directly defined on the sphere. Notably, the Sph-IoU calculation moves the Sph-BBs to the equator along the longitude, making the intersection calculation more convenient as the adjusted Sph-BBs become rectangular. However, moving the centers of Sph-BBs can improve IoU calculation but also alter the great-circle distance between two centers, which should be preserved. Therefore, Zhao~\etal\cite{Zhao2021UnbiasedIF} and Cao~\etal\cite{Cao2022FieldofViewIF} calculated unbiased spherical IoUs based on the intersection area between overlapping spherical rectangles on the sphere (referred to as Field of View BBs or FoV-BBs, as shown in Fig.~\ref{fig:objection_detection_BBs}(b)). Inspired by objects with arbitrary orientations in ODIs, rotated Field of View BBs (RFoV-BBs)\cite{Xu2022PANDORAAP} are proposed based on FoV-BBs, incorporating the angle of rotation of the tangent plane of the RFoV-BBs. Furthermore, similar to Sph-BBs, Sph2Pob\cite{Liu2023Sph2PobBO} is proposed to move the arc between the two FoV-BBs' centers to the equator, making it easier to map to planar-oriented boxes. Based on Sph2Pob, a differentiable IoU, Sph2Pob-IoU, is utilized to approximate spherical IoU with IoU for planar-oriented boxes.

Differently, GLDL~\cite{Xu2023GaussianLD} provides a novel direction by modeling FoV-BBs using Gaussian distributions and calculating the K-L divergence between the two distributions as the sample selection strategy. This distribution-based selection strategy can alleviate the drawback of the IoU threshold-based strategy, which is weak to the uneven scales of objects from different categories.

\noindent\textbf{Discussion:} Through the aforementioned analysis, it is evident that in ODI object detection, the key challenge lies in determining the BBs of objects and computing the IoUs between different BBs, compared to object detection in perspective images. Due to the spherical imaging nature of ODIs, conventional rectangular bounding boxes are often inadequate, particularly for distorted objects in ERP format panoramas. Consequently, several methods have been proposed to introduce spherical BBs for improved representation of objects. Specifically, the ability of ODIs to freely rotate enables the movement of spherical BBs, facilitating the computation of IoU between them. The approach presented in GLDL, utilizing Gaussian distribution to establish differences between various bounding boxes for better positive sample selection, offers an intriguing direction on how to effectively leverage spherical geometry to explore the relationships between different bounding boxes. Concurrently, recent advancements in planar generative models in planar object detection, such as DiffusionDet~\cite{Chen2022DiffusionDetDM}, have spurred an exploration into generating spherical bounding boxes via more flexible ways.

\vspace{-10pt}
\subsubsection{Segmentation}
\vspace{-5pt}
\label{sec5.2.2}
Learning-based panoramic semantic segmentation has attracted significant attention due to the comprehensive information it provides about the surrounding space, which is crucial for scene understanding. However, this field encounters several practical challenges, including distortions in planar projections, object deformations, semantic complexity in wide-FoV scenes, and limited labeled data. In the following sections, we will delve into the development of this field by analyzing some representative works.

Initially, people mostly cover the wide Field of View (FoV) up to 360 degrees by arranging multiple narrow-FoV perspective cameras~\cite{Deng2017CNNBS} or connecting fisheye cameras with obvious lens distortion~\cite{DengYLLHW20}. However, the use of multiple camera systems often led to large processing latency and required complex calibration tasks. In contrast, recent panoramic semantic segmentation methods typically utilize a single panorama as input, following two main approaches: 1) Knowledge transfer from perspective domain to panoramic domain; 2) Direct prediction of panoramic semantic segmentation results while considering the 3D spherical properties.
\begin{figure}[!t]
  \centering
  \includegraphics[width=0.9\linewidth]{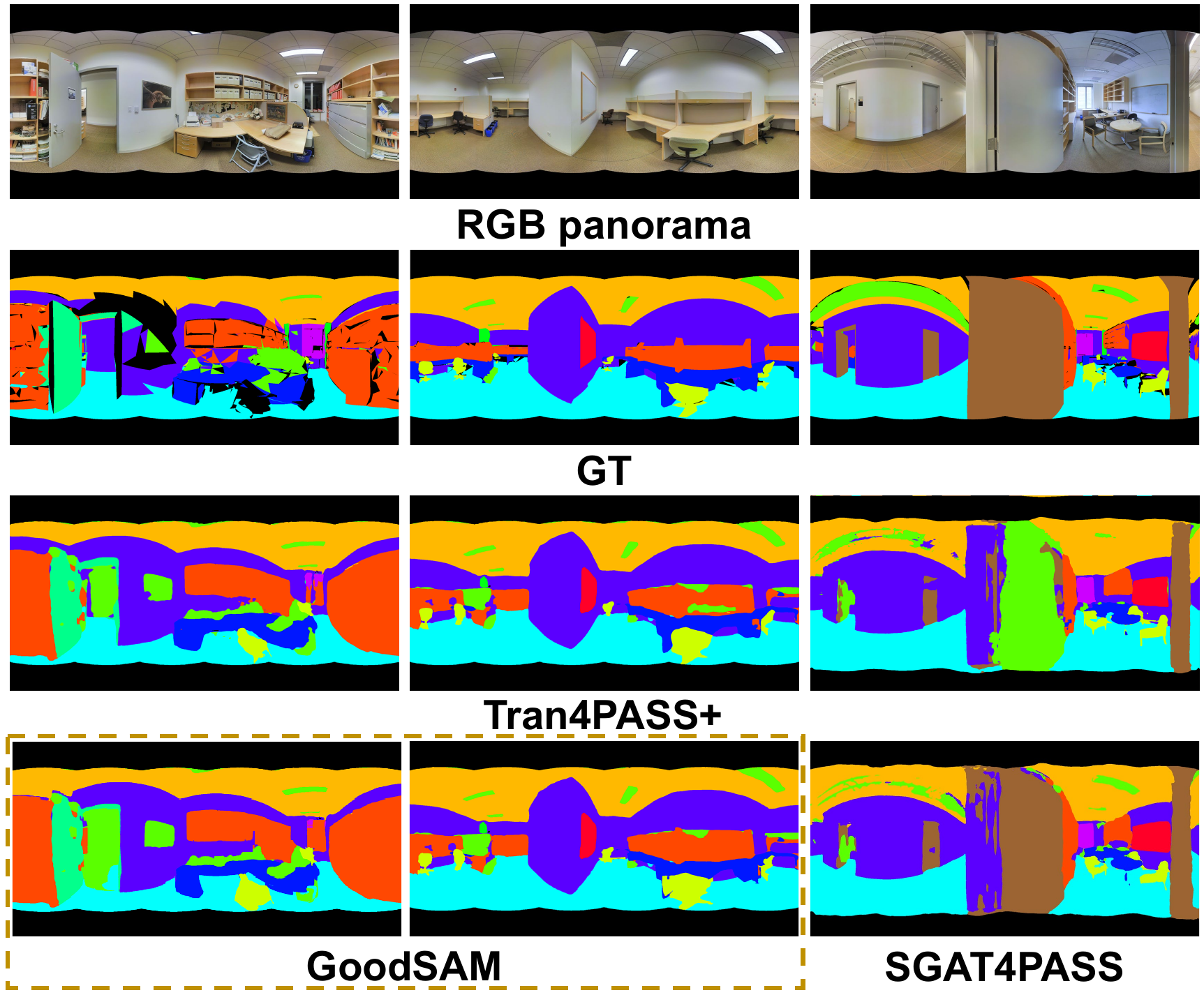}
  \vspace{-8pt}
  \caption{Performance of SoTA panoramic segmentation methods: Trans4PASS~\cite{Zhang2022BendingRD}, SGAT4PASS~\cite{Li2023SGAT4PASSSG}, and GoodSAM~\cite{zhang2024goodsam}, as reported in their original papers.} 
    \label{fig:segmentation}
    \vspace{-15pt}
\end{figure}

PASS~\cite{yang2019can},~\cite{YangHBRW20} proposes the first panoramic semantic segmentation framework based on perspective knowledge. Specifically, PASS utilizes a pre-trained segmentation network from pinhole imagery to omnidirectional imagery, thereby bypassing the need for extensive pixel-exact annotations for panoramas. Building on PASS, Yang~\etal~\cite{Yang2019DSPASSDP} further enhanced the knowledge transfer performance from the 2D domain to the panoramic domain. They introduced attention-based lateral connections to improve sensitivity to spatial details. Both PASS and DS-PASS divide a panorama into segments for semantic predictions using pre-trained 2D segmenters, which are then fused back into a complete panorama. To mitigate computational complexity and capture global scenes, Yang~\etal~\cite{yang2020omnisupervised} employed data distillation between labeled pinhole images and unlabeled panoramas. This approach learns a teacher model from labeled pinhole images and generates semantic labels for training the student network. The data distillation strategy is also utilized in~\cite{Yang2021IsCC},~\cite{Yang2021CapturingOC}, where hand-crafted feature extraction modules are proposed to capture contextual information of wide-FoV panoramas.

Other researchers have explored the concept of unsupervised domain adaptation for panoramic semantic segmentation. A prominent work in this area is Dense-PASS~\cite{ma2021densepass},~\cite{Zhang2021TransferBT}, which introduces a generic pinhole-to-panoramic domain adaptation (P2PDA) pipeline. This pipeline adapts semantic segmentation networks from the label-rich source domain of pinhole camera images to the target domain of unlabeled panoramas. To achieve better cross-domain alignment, Dense-PASS specifically designs several domain adaptation modules, which can be activated individually or jointly within the P2PDA framework. In addition to the gap caused by different FoVs, spherical distortion also presents a significant domain gap. Addressing this, Trans4PASS~\cite{Zhang2022BendingRD} and Trans4PASS+~\cite{Zhang2022BehindED} propose a distortion-aware transformer with several deformable components as the semantic segmentation network. Meanwhile, Inspired by~\cite{Zhang2021PrototypicalPL}, Trans4PASS employs a semantic prototypes-based domain adaptation strategy to generate pseudo labels for unlabeled panoramas. DPPASS~\cite{zheng2023both} leverages the geometric correspondence between two types of projections,~\ie, less-distorted tangent projection and ERP, for cross-projection domain adaptation. Besides, DPPASS performs cross-style domain adaptation by transforming pinhole images into pseudo ERP and TP formats and aligning the pinhole domain with the panoramic domain. Moreover, Zheng~\etal~\cite{Zheng2023LookAT} exploited the non-uniform sampling density of ERP projection, quantified distortion, and proposed minimizing the receptive field of the segmentation backbone to address the distortion problem. Recently, Zheng~\etal further investigated a source-free unsupervised panoramic segmentation method, 360SFUDA~\cite{zheng2024semantics}, which leverages the semantic, distortion, and style similarity between ERP slices and tangent patches to achieve knowledge transfer, enabling effective pinhole-to-panoramic adaptation. With the rise of large vision models, Zhang~\etal~\cite{zhang2024goodsam} evaluated the SAM model~\cite{kirillov2023segment} on panoramas and introduced GoodSAM, which enhances SAM's zero-shot segmentation capabilities for panoramic semantic segmentation. By integrating the instance masks from SAM and logits from a teacher-assistant model, GoodSAM generates high-quality guidance to achieve unsupervised panoramic semantic segmentation.

In contrast, supervised panoramic semantic segmentation research based on spherical geometry remains an open field. Zhang~\etal~\cite{zhang2019orientation} converted the given spherical panorama into an unfolded icosahedron mesh and utilized a customized orientation-aware CNN for segmentation prediction. Li~\etal~\cite{Li2023SGAT4PASSSG} addressed the high sensitivity of existing 2D segmentation models to 3D disturbances and proposed the spherical geometry-aware framework. Specifically, they employed random rotations to the given panoramas and incorporated spherical geometry into deformable patch embedding~\cite{Zhang2022BendingRD} to construct the transformer.
\begin{table}[!t]
    \centering
    \caption{Quantitative comparison of the methods for UDA-based panoramic annular semantic segmentation task on DensePASS dataset.
    }
    \vspace{-5pt}
    \label{tab:segmentation/comparison-segmentation-uda}
    \resizebox{0.95\linewidth}{!}{ 
    \begin{tabular}{c|c|c}
    \toprule
    Method  & $\#$ Para. (M) & mIoU ($\%$)\\
    \midrule
    PASS~\cite{YangHBRW20} & $-$ & 23.66 \\
    Yang~\etal~\cite{yang2020omnisupervised}& $-$ & 23.66 \\
    Dense-PASS~\cite{Zhang2021TransferBT}  & $-$ & 43.02 \\
    Trans4PASS-Small~\cite{Zhang2022BendingRD}& 24.98&55.22\\
    DPPASS-Small~\cite{zheng2023both}& 25.40&56.28\\
    DATR-Small~\cite{Zheng2023LookAT}& 25.76&56.81\\
    GoodSAM-Small~\cite{zhang2024goodsam}&25.40&60.56\\
    \bottomrule
    \end{tabular}}
    \vspace{-5pt}
\end{table}
\begin{table}[!t]
    \centering
    \caption{Quantitative comparison of the methods for supervised panoramic semantic segmentation task on Stanford2D3D dataset.
    }
    \vspace{-5pt}
    \label{tab:segmentation/comparison-segmentation-sup}
    \resizebox{0.95\linewidth}{!}{ 
    \begin{tabular}{c|c|c}
    \toprule
    Method   &  Avg mIoU ($\%$) & F1 mIoU \\
    \midrule
    Tangent (ResNet-101)~\cite{eder2020tangent}&45.6&-\\
    PanoFormer~\cite{shen2022panoformer} & 48.9 & -\\
    HoHoNet (ResNet-101)~\cite{sun2021hohonet} & 52.0&53.9\\
    Trans4PASS-Small~\cite{Zhang2022BendingRD}&52.1&53.3\\
    SGAT4PASS~\cite{Li2023SGAT4PASSSG}& 55.3&56.4\\
    \bottomrule
    \end{tabular}}
    \vspace{-15pt}
\end{table}

\noindent\textbf{Discussion:} Tab.~\ref{tab:segmentation/comparison-segmentation-uda} and Tab.~\ref{tab:segmentation/comparison-segmentation-sup} show state-of-the-art results of unsupervised domain adaptation (UDA)-based panoramic semantic segmentation methods and supervised panoramic semantic segmentation methods, respectively. We consider the DensePASS~\cite{ma2021densepass} and Stanford2D3D~\cite{armeni2017joint} datasets. Especially, for the Stanford2D3D dataset, we listed the performance of official fold 1 and the average performance of all three official folds. To our best knowledge, GoodSAM~\cite{zhang2024goodsam} has achieved the most promising results among the UDA-based methods thus far, while SGAT4PASS~\cite{Li2023SGAT4PASSSG} has notably enhanced the performance of supervised semantic segmentation. Furthermore, to effectively demonstrate the performance of previous SoTA methods, we provide the visual results of Trans4PASS+\cite{Zhang2022BehindED}, SGAT4PASS\cite{Li2023SGAT4PASSSG}, and GoodSAM~\cite{zhang2024goodsam} on Stanford2D3D dataset (See Fig.~\ref{fig:segmentation}).

Through the previous analysis, it can be observed that the current focus remains on unsupervised domain adaptation (UDA) for panoramic semantic segmentation. This preference is primarily due to the ease of obtaining accurate annotations for traditional pinhole camera images, which facilitates data-driven learning methods. However, existing UDA-based approaches overlook the spherical geometry of panoramas, such as pitch and roll rotations. Given that 3D robustness presents a significant gap between pinhole and panoramic images, enhancing the 3D robustness of segmentation models in the UDA process is a crucial area for future research. In particular, it is worth noting that, in most current UDA-based methods, the 'panoramic images' they use do not strictly adhere to the definition of panoramas but are \textit{panoramic annular images} (as mentioned in PASS~\cite{yang2019can,YangHBRW20,Yang2019DSPASSDP}). The difference is that panoramic annular images do not have seriously distorted poles like panoramas, so the impact of distortion on UDA-based methods still deserves further exploration.

Additionally, panoramic panoptic segmentation, aiming to simultaneously predict semantic and object instance information, is of significant value for autonomous driving. Nevertheless, this area of research remains relatively unexplored. Mei~\etal~\cite{mei2022waymo} collected a large-scale panoramic video panoptic segmentation dataset with dense, high-quality labels. Similar to unsupervised panoramic semantic segmentation, Jaus~\etal~\cite{jaus2021panoramic,jaus2023panoramic} improved the performance of panoramic panoptic segmentation by leveraging models trained on richly labeled pinhole images. Recent work~\cite{kinzig2024panoptic} proposed to predict segmentation results from multiple less-distorted NFOV regions and stitch them into a panoramic view. However, this method suffers from limited real-time capability. Panoramic panoptic segmentation requires finer-grained scene understanding, making it promising to adapt the feature extraction capabilities of planar vision models to panoramas. Additionally, collecting high-quality real-world annotations can support the development of sphere-specific learning algorithms to advance this field.

\vspace{-5pt}
\subsubsection{Saliency Prediction}
\label{sec5.2.3}
\begin{figure}[!t]
  \centering
  \includegraphics[width=\linewidth]{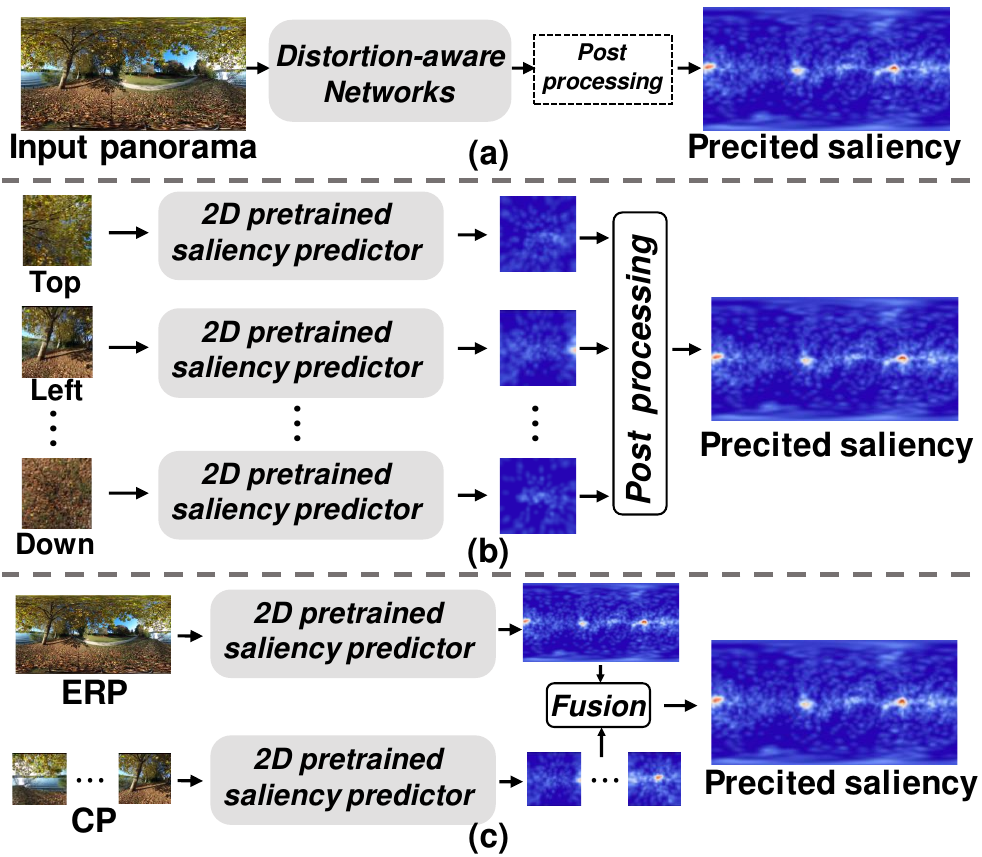}
  \vspace{-10pt}
  \caption{Typical pipelines of ODI saliency prediction: (a) ERP-based;(b) CP-based;(c) ERP with CP-based.} 
    \label{fig:saliency}
    \vspace{-15pt}
\end{figure}

\vspace{-5pt}
As ODIs/ODVs can offer an immersive experience to observers within Head-Mounted Displays (HMDs), it is crucial to predict the most visually significant or attention-grabbing areas in a 360$^\circ$ scene. This not only aids in designing the user interface and eye tracking for future VR systems but also contributes to understanding the visual behaviors of users in virtual environments. In the literature, most methods have focused on two projection formats: ERP and CP. They either use ERP or CP individually as input or combine both as joint inputs. Tab.~\ref{tab:saliency_prediction/methods_saliency_prediction} presents an overview of the reviewed approaches and Fig.~\ref{fig:saliency} concludes some representative types of ODI saliency prediction frameworks. 

\noindent\textbf{ODI Saliency Prediction.} In Fig.~\ref{fig:saliency}(a), with a single panorama as the input, Zhu~\etal~\cite{Zhu2020RanspRA} proposed a feature filtering mechanism, which ranks the extracted features and adaptively selects desired feature channels according to the ranking scores, avoiding the redundant information in the features extracted by deep convolutional layers. 
Some authors employ diverse convolution filters to enhance the feature representation capability. Zhu~\etal~\cite{Zhu2022DecoupledDG} designed a decoupled dynamic group equivariant filter based on group equivariant convolution filter (p4-convolution)~\cite{cohen2016group} to extract content-adaptive features from ERP input. The authors decoupled p4-convolution into spatial and channel dynamic filters which are content-agnostic and efficient for better saliency prediction. Zhu~\etal~\cite{Zhu2021ALS} proposed a light-weight saliency model using the dynamic convolution~\cite{Chen2019DynamicCA} and aggregated multiple parallel convolution kernels with the attention weights for extract powerful features from ERP input. Besides, Zhu~\etal~\cite{Zhu2020ThePO} proposed a novel approach in which they employed spherical harmonics to represent the panorama. They utilized this representation to extract features across various frequency bands and orientations, enabling the estimation of saliency. Ma~\etal~\cite{Ho2016GenerativeAI} built the ODI saliency model, SalGAIL, via generative adversarial imitation learning (GAIL)~\cite{Xu2019SaliencyPO}. SalGAIL predicts the head fixations of users and then estimates the saliency through the predicted head fixations. Based on the consistency between head fixations and subjects in ODIs, SalGAIL applies the deep reinforcement learning to predict the head fixations of one subject and use the GAIL to learn the reward of DRL. Recently, MFFPANet~\cite{Chen2024MultiStageSO} takes an ERP as input to predict object-level semantic images, which are taken as joint inputs with the panorama. In MFFPANet, a transformer-based network is proposed to aggregate contextual long-range information.
\begin{table*}[!t]
    \centering
    \caption{ODI and ODV Saliency prediction methods. EM and HM mean eye and head movement.}
    \vspace{-8pt}
    \label{tab:saliency_prediction/methods_saliency_prediction}
    \resizebox{0.92\linewidth}{!}{ 
    \begin{tabular}{c|c|c|c|c|c|c}
    \toprule
    &Reference  & Publication &Category & Input & Technology & Contribution\\
    \midrule
    \multirow{14}*{Image}
    &Monroy~\etal~\cite{Monroy2017SalNet360SM}&SPIC'18&HM$\&$EM& CP& CNN & Transfer learning\\&Dedhia~\etal~\cite{Dedhia2019SaliencyPF}&ICASSP'19&HM$\&$EM&ERP$\&$CP&CNN&Transfer learning\\
    &Zhu~\etal~\cite{Zhu2020RanspRA}&ICME'20& HM&ERP& CNN $\&$ Attention & Specific attention modules\\
    &Djemai~\etal~\cite{Djemai2020Extending2S}&ICME'20&HM&CP&CNN&Rotations on input\\
    &Chen~\etal\cite{Chen2020SalBiNet360SP}&VR'20&HM$\&$EM&ERP$\&$CP&CNN&Bi-projection fusion\\
    &Lv~\etal~\cite{Lv2020SalGCNSP}&ACM MM'20&HM$\&$EM&spherical graph&GCN&Graph saliency prediction\\
    &Dai~\etal~\cite{Dai2020DilatedCN}&ICASSP'20&HM$\&$EM&CP&CNN&Dilated convolution\\
    &Zhu~\etal~\cite{Zhu2020ThePO}&TMM'20&HM$\&$EM&ERP& CNN & Spherical harmonics\\
    &Zhu~\etal~\cite{Zhu2021ALS}&ICME'21&EM&ERP&CNN $\&$ Attention&Dynamic convolution \\
    &Chen~\etal~\cite{Chen2022IntraAI}&TVSVT'22&EM&CP&GCN&Intra- and Inter-graph inference\\
    &Chen~\etal~\cite{Chen2024MultiStageSO}&TETCI'24&HM$\&$EM&ERP& CNN & Feature fusion\\
    \midrule
    \multirow{7}*{Video}&Nguyen~\etal~\cite{Nguyen2018YourAI}  & ACM MM'18& HM &  ERP & CNN $\&$ LSTM & Transfer learning\\
    &Chen~\etal~\cite{cheng2018cube}&CVPR'18&EM& CP & CNN$\&$ConvLSTM& Cube padding\\
    &Zhang~\etal~\cite{Zhang2018SaliencyDI}&ECCV'18&EM&ERP&CNN&spherical CNN\\
    &Xu~\etal~\cite{Xu2019PredictingHM}&TPAMI'19&HM&ERP$\&$TP &CNN$\&$LSTM&Deep reinforcement learning\\
    &Qiao~\etal~\cite{Qiao2021ViewportDependentSP}&TMM'21&HM$\&$EM&ERP$\&$TP&CNN$\&$ConvLSTM&Multi-task learning\\
    &Zhu~\etal~\cite{Zhu2021ViewingBS}&TCSVT'21&HM$\&$EM&ERP$\&$TP &CNN$\&$GCN&Visual attention\\
    \bottomrule
    \end{tabular}}
    \vspace{-15pt}
\end{table*}
Due to the lack of large-scale ODI/ODV datasets with accurate pixel-accurate annotations, many methods are exploring how to leverage more abundant perspective image datasets to aid in the saliency prediction of ODIs/ODVs. Considering the gap between ERP images and perspective images,~\eg, spherical distortion, some methods proposed to use less-distorted cubemap projection patches as input, as illustrated in Fig.~\ref{fig:saliency}(b). As the first attempt of the pre-trained 2D saliency model on ODI saliency prediction, SalNet360~\cite{Monroy2017SalNet360SM} splits an ERP into a set of six CP patches as the input via the geometric relationships (See Fig.~\ref{fig:imaging}). Then SalNet360 combines predicted saliency maps and per-pixel spherical coordinates of these patches to output a resulting saliency map in ERP format. As CP patches inherently exhibit distortion, some methods are proposed to mitigate this effect. Dai~\etal~\cite{Dai2020DilatedCN} proposed the use of dilated convolutional layers to extract features from CP patches, providing a method to address distortion in CP slices. Djemai~\etal~\cite{Djemai2020Extending2S} provided a pipeline that random rotates the content of each CP patch before applying 2D saliency methods to CP patches. This can make each CP patch provide a larger context and enhance the performance of 2D saliency methods. In contrast, SalReGCN360~\cite{Chen2022IntraAI} establishes spatial and semantic correlations among the set of CP patches, enabling the saliency model to perceive global information. Specifically, SalReGCN360 introduces a graph convolutional network (GCN) to enhance spatial and semantic associations among CP features.

Some works take the panorama and CP patches as the joint inputs, as shown in Fig.~\ref{fig:saliency}(c). They~\cite{Dedhia2019SaliencyPF},~\cite{Chen2020SalBiNet360SP} extend the 2D saliency predictor to omnidirectional imagery and leverage the fusion of saliency results from ERP and CP to generate the final results.~\cite{Dedhia2019SaliencyPF} employs the equator bias map to adjust the fused result and mimic the viewers’ behavior, while SalBiNet360~\cite{Chen2020SalBiNet360SP} builds a two-branch local-global bifurcated deep network to process ERP and CP input separately.

In particular, some works explore the potential of other input representations.~\cite{Suzuki2018SaliencyME} applies a 2D saliency model on multiple viewports, re-projects all the viewports' results into an ERP format saliency map, and emphasizes the gaze at the equator with equator bias map. Differently from the aforementioned works,~\cite{Lv2020SalGCNSP},~\cite{Yang2021SalGFCNGB} propose to represent the ODIs as non-Euclidean spherical graphs and build the graph convolution networks to predict the saliency maps. SalGCN~\cite{Lv2020SalGCNSP} is the pioneer representative work with the spherical graph signals, which are constructed based on the geodesic icosahedral pixelation. Then, using a U-Net structured graph convolutional network, a saliency map in spherical graph format is predicted. Then, through a designed spherical crown-based interpolation module, the spherical graph is transformed into ERP format. While SalGFCN~\cite{Yang2021SalGFCNGB} is also based on a spherical graph, unlike SalGCN, the graph construction in SalGFCN is based on Spherical Fibonacci Mapping (SFM)~\cite{keinert2015spherical}. Additionally, the prediction network of SalGFCN is composed of a residual U-Net architecture incorporating dilated graph convolutions and an attention mechanism in the bottleneck. The process of transforming the graph into ERP format is based on distance-based interpolation.

\noindent\textbf{ODV Saliency Prediction.} For the saliency prediction in ODVs, the key points are accurate saliency prediction for each frame and the temporal coherence of the viewing process. As videos with dynamic contents are widely used in real applications, deep ODV saliency prediction has received more attention in the community. Nguyen \etal~\cite{Nguyen2018YourAI} proposed a representative transfer learning framework that fine-tunes existing 2D saliency predictors for ERP input and employs a structure-aware prior filter to enhance the predicted results, targeting specific areas such as the four corners of an input panorama. For cubemap projection format input, customized Cube Padding~\cite{cheng2018cube} is an influenced method, whose proposed padding operation can mitigate patch boundaries among CP patches, and be generally applicable to almost all existing 2D CNN architectures. Moreover, they proposed a spatial-temporal network consisting of a static~\cite{Zhou2016LearningDF} model and a ConvLSTM~\cite{shi2015convolutional} module. Similar to~\cite{Suzuki2018SaliencyME}, a viewport saliency prediction model is proposed in~\cite{Qiao2021ViewportDependentSP} which first studies human attention to detect the desired viewports of the ODV and then predict the fixations based on the viewport content. One more representative is proposed by~\cite{Zhang2018SaliencyDI}, in which the convolution kernel is defined on a spherical crown and the convolution operation corresponds to the rotation of kernel on the sphere. Considering the common planar ERP format, Zhang \etal~\cite{Zhang2018SaliencyDI} re-sampled the kernel based on the position of the sampled patches on ERP. There also exist some works based on novel learning strategies. Xu \etal~\cite{Xu2019PredictingHM} developed the saliency prediction network of head movement (HM) based on deep reinforcement learning (DRL). The proposed DRL-based head movement prediction approach owns offline and online versions. In offline version, multiple DRL workflows determines potential HM positions at each panoramic frame and generate a heat map of the potential HM positions. In online version, the DRL model will estimate the next HM position of one subject according to the currently observed HM position. Zhu \etal~\cite{Zhu2021ViewingBS} proposed a graph-based CNN model to estimate the fraction of the visual saliency via Markov Chains. The edge weights of the chains represent the characteristics of viewing behaviors, and the nodes are feature vectors from the spatial-temporal units.

\vspace{-10pt}
\subsection{3D Geometry and Motion Estimation}
\vspace{-5pt}
\label{sec5.3}

\subsubsection{Depth Estimation}
\vspace{-5pt}
\label{sec5.3.1}

\begin{figure}[!t]
  \centering
  \includegraphics[width=0.92\linewidth]{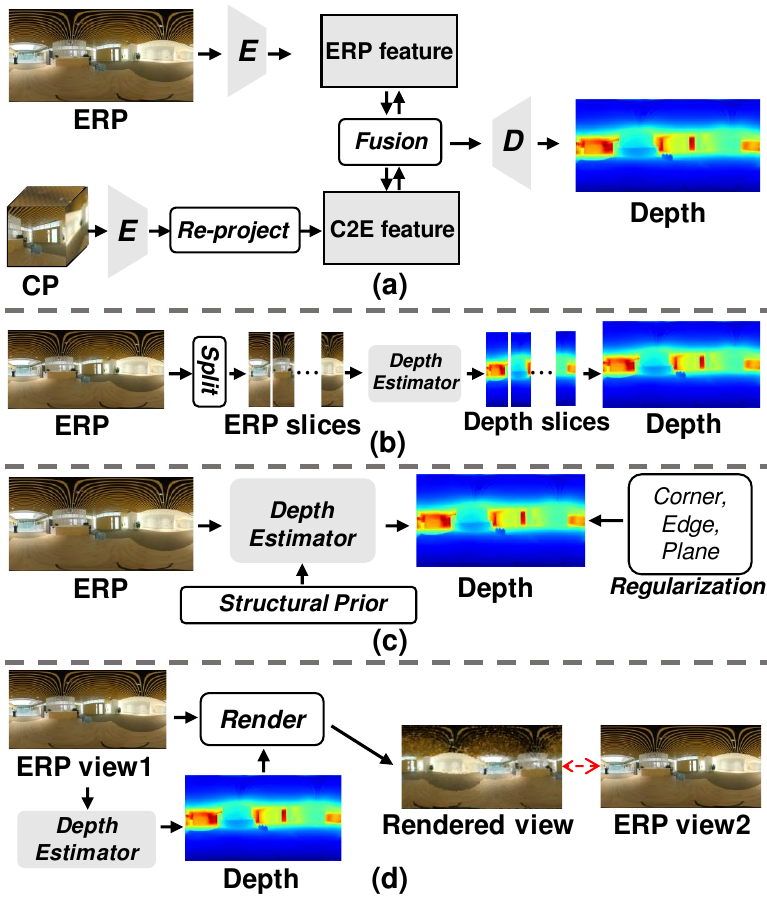}
  \vspace{-8pt}
  \caption{Representative monocular depth estimation methods: (a) Multi-projection input; (b) 2D representations (~\eg, vertical slices of ERP); (c) Extra geometric prior knowledge; (d) View synthesis.} 
    \label{fig:Monocular depth estimation}
    \vspace{-10pt}
\end{figure}

The commonly used ERP format of ODIs exhibits horizontal distortion. Consequently, ERP images often exhibit increased curves compared to perspective images. It would also result in neighboring points on the sphere becoming more distant. Considering that the number of monocular methods for ODIs is larger than stereophonic methods by a great deal, we mainly focus on monocular methods in this subsection. For the task of monocular depth estimation task, 2D methods typically apply regular convolutional kernels (\eg, $3 \times 3$) to extract geometric information. However, the direct application of planar designs struggles to capture effective geometric structures due to the increased curves and distant neighbors, leading to sub-optimal results. To address this issue, recent trends can be grouped into five directions: (i) Tailored networks, such as distortion-aware convolution filters~\cite{Tateno2018DistortionAwareCF} and attentions~\cite{yun2023egformer}; (ii) Exploring more projection types of ODIs with less distortion~\cite{wang2020bifuse},~\cite{feng2022360},~\cite{li2022omnifusion} (Fig.~\ref{fig:Monocular depth estimation}(a)); (iii) Developing 2D representations specifically for ERP~\cite{Pintore2021SliceNetDD} (Fig.~\ref{fig:Monocular depth estimation}(b)); (iv) Leveraging inherent geometric priors contained in ODI~\cite{eder2019pano},~\cite{jin2020geometric} (Fig.~\ref{fig:Monocular depth estimation}(c)); (v) Employing multiple views of ODIs~\cite{zioulis2019spherical} (Fig.~\ref{fig:Monocular depth estimation}(d)). We further provide a detailed accuracy comparison of these methods in Tab.~\ref{tab:deph/comparison-depth}, which will be discussed subsequently. Finally, we introduce the ODI depth completion with additional sparse depth input, and stereo matching within two paired ODIs.

\noindent \textbf{Tailored networks:} This category can be divided into two subsets: specialized CNN design and attention design. In terms of CNN design, the main emphasis is on the shape and sampling positions of the CNN kernel. To mitigate the effects of stretch distortion, Zioulis \etal~\cite{zioulis2018omnidepth} building on \cite{Su2017LearningSC}, transformed regular square convolution filters into row-wise rectangles and adjusted filter sizes across latitudes to compensate for distortions near the poles. Tateno \etal~\cite{Tateno2018DistortionAwareCF} introduced a deformable convolution filter that samples the pixel grids on the tangent planes and then projects them back to the unit sphere coordinates. More recently, Zhuang \etal~\cite{zhuang2022acdnet} developed an approach combining a series of dilated convolutions with varying dilation rates, which adaptively enlarges the receptive field in the ERP images. In the realm of attention design, PanoFormer~\cite{shen2022panoformer} introduces pixel-wise deformable attention,  echoing the principles of~\cite{coors2018spherenet}. However, PanoFormer further incorporates a learnable token flow to ensure that the sampled positions of the attention mechanism align closely with the geometric structures in the scene. EGFormer~\cite{yun2023egformer} introduces a relative position embedding that calculates the distances between pixels in spherical coordinates, which is then utilized to re-weight the attention scores. 
% Building upon the Swin transformer, PanoSwin~\cite{ling2023panoswin} designs window shift operations in accordance with spherical rotation principles. Additionally, PanoSwin introduces a pitch attention module to reorient regions with significant distortion towards the equatorial regions, where distortion is less pronounced.

% In comparison, Pintore \etal~\cite{Pintore2021SliceNetDD} proposed a framework, named SliceNet, with regular convolution filters to work on the ERP directly. SliceNet reduces the input tensor only along the vertical direction to collect a sequence of vertical slices and adopts an LSTM~\cite{shi2015convolutional} network to recover the long- and short-term spatial relationships among slices.

\begin{table*}[!t]
    \centering
    \caption{Quantitative comparison for monocular ODI depth estimation task on two popular datasets.
    }
      \vspace{-8pt}
    \label{tab:deph/comparison-depth}
    \resizebox{1\linewidth}{!}{ 
    \begin{tabular}{c|c|c|c|c|c|c|c|c|c|c}
    \toprule
    Datasets&Method & Publication & Backbone & Input Format & Abs Rel $\downarrow$& Sq Rel $\downarrow$& RMSE $\downarrow$ &$\delta_1$ $\uparrow$ & $\delta_2$ $\uparrow$ & $\delta_3$ $\uparrow$\\
    \midrule
    \multirow{8}*{Stanford2D3D} 
&BiFuse~\cite{wang2020bifuse}&CVPR'20&Res-50&ERP+CP&0.1209&$-$&0.4142&0.8660&0.9580&0.9860\\
&UniFuse~\cite{jiang2021unifuse}&RAL'21&ResNet-18&ERP+CP&0.1114&$-$&0.3691&0.8711&0.9664&0.9882\\
&HoHoNet~\cite{sun2021hohonet}&CVPR'21&ResNet-50&ERP&0.1014&$-$&0.3834&0.9054&0.9693&0.9886\\
&OmniFusion~\cite{li2022omnifusion}&CVPR'22&ResNet-34&TP&0.0950&0.0491&0.3474&0.8988&0.9769&0.9924\\  &PanoFormer~\cite{shen2022panoformer}&ECCV'22&$-$&ERP& 0.1131&0.0723  & 0.3557& 0.8808& 0.9623&0.9855\\
&PanelNet~\cite{yu2023panelnet}&CVPR'23&ResNet-34&ERP&$-$&0.0829 &0.2933 &0.9242 & 0.9796&0.9915\\
&HRDFuse~\cite{ai2023hrdfuse} &CVPR'23&ResNet-34&ERP+TP&0.0935&0.0508&0.3106&0.9140 & 0.9798&0.9927\\
 &S2Net~\cite{li2023mathcal} & RAL'23&Swin-Base &ERP+HEALPix& 0.0903 & $-$ & 0.3383 & 0.9191 & 0.9782 & 0.9912 \\
      \midrule
    \multirow{8}*{Matterport3D}
&BiFuse~\cite{wang2020bifuse}&CVPR'20&ResNet-50&ERP+CP&0.2048&$-$&0.6259&0.8452&0.9319&0.9632\\
&UniFuse~\cite{jiang2021unifuse}&RAL'21&ResNet-18&ERP+CP&0.1063&$-$&0.4941&0.8897&0.9623&0.9831\\
&HoHoNet~\cite{sun2021hohonet}&CVPR'21&ResNet-50&ERP&0.1488&$-$&0.5138&0.8786&0.9519&0.9771\\
    &OmniFusion~\cite{li2022omnifusion} &CVPR'22&ResNet-34&TP&0.1007&0.0969&0.4435&0.9143&0.9666&0.9844\\
&PanoFormer~\cite{shen2022panoformer}&ECCV'22&$-$&ERP& 0.0904&0.0764& 0.4470& 0.8816& 0.9661& 0.9878\\
&HRDFuse~\cite{ai2023hrdfuse} &CVPR'23&ResNet-34&ERP+TP&0.0967&0.0936&0.4433&0.9162&0.9669&0.9844\\
&PanelNet~\cite{yu2023panelnet}&CVPR'23&ResNet-34 & ERP & 0.1150&$-$& 0.4528&0.9123&0.9703 &0.9856 \\
 &S2Net~\cite{li2023mathcal} & RAL'23&Swin-Base &ERP+HEALPix& 0.0865 & $-$ & 0.4052 & 0.9264 & 0.9768 & 0.9911 \\

    % \midrule
    % \multirow{6}*{360D} &BiFuse~\cite{wang2020bifuse}&CVPR'20&ResNet-50&0.0615&$-$&0.2440&0.9699 &0.9927 &0.9969\\
    % &UniFuse~\cite{jiang2021unifuse}&RAL'21&ResNet-18& 0.0466 &- &0.1968&0.9835 &0.9965& 0.9987\\
    % &OmniFusion~\cite{li2022omnifusion}&CVPR'22&ResNet-34 &0.0430 & 0.0114 & 0.1808& 0.9859 & 0.9969 & 0.9989\\
    % &PanoFormer~\cite{shen2022panoformer}&ECCV'22&$-$&0.0501  & $-$& 0.1492  &0.9867&0.9975&0.9991 \\
    % &HRDFuse~\cite{ai2023hrdfuse} &CVPR'23&ResNet-34&0.0358&0.0100&0.1555&0.9894&0.9973&0.9990\\
    %  &S2Net~\cite{li2023mathcal} & RAL'23&Swin-Base & 0.0357 & $-$ & 0.1818 & 0.9911 & 0.9971 & 0.9987 \\
    \bottomrule
    \end{tabular}}
    \vspace{-15pt}
\end{table*}

\noindent \textbf{Different Projection Formats:} Efforts have been made to mitigate the distortion in ERP by exploring other projection formats with less distortion, \eg, Cubemap Projection (CP), and tangent projection, as shown in Fig.~\ref{fig:Monocular depth estimation}(a). BiFuse~\cite{wang2020bifuse}, a notable study, introduces a dual-branch architecture in which one branch processes the ERP input, while the other extracts features from CP to mimic the human peripheral and foveal vision. A fusion model is then proposed to process the semantic and geometric information from both branches. Drawing inspiration from BiFuse, UniFuse~\cite{jiang2021unifuse} proposes a more efficient fusion module to combine the features from two branches, in which the CP features are passed exclusively to the ERP branch during the decoding phase. To enhance the global context extraction, GLPanoDepth~\cite{bai2022glpanodepth} transforms the ERP input into a collection of CP images and employs a Vision Transformer (ViT) model to learn long-range dependencies. Given that tangent projection yields less distortion than CP, 360MonoDepth~\cite{ReyArea2022360MonoDepthH3} employs state-of-the-art 2D depth estimation models~\cite{ranftl2021vision} to predict depths from tangent images, which are then readjusted and merged to ERP depths with alignment and blending techniques. Nevertheless, directly re-projecting tangent images back to ERP can lead to overlapping and discontinuities. To ensure global consistency, Peng~\etal~\cite{peng2023high} proposed to employ a pre-computed ERP depth map from the pre-trained depth estimator to support other patch-wise depth predictions' registrations. Additionally, OmniFusion~\cite{li2022omnifusion} introduces additional 3D geometric embeddings to lessen the discrepancy in patch-wise features and aggregate such information with a transformer. Considering that tangent projection is limited to local regions, HRDFuse~\cite{ai2023hrdfuse} unveils a dual-branch strategy, comprising an ERP branch and a TP branch. The integration of the two branches is facilitated by learnable feature alignment, which offers benefits in both efficiency and accuracy compared to geometric fusion. Recently, S2Net~\cite{li2023mathcal} introduces to project ERP features onto the sphere with HEALPix~\cite{Gorski2005HEALPixAF}, which achieves uniform sampling on the sphere. Especially, HEALPix transforms an ERP into a point set and S2Net processes the point set with a cross-attention block.

\noindent \textbf{2D Representations:} This approach pertains to the ERP representation which is vertically captured or aligned with gravity. To exploit the gravity-aligned features in ERP, Pintore~\etal~\cite{Pintore2021SliceNetDD} developed a framework called SliceNet, which deconstructs the ERP into vertical slices and reduces the input tensor along the vertical (gravity) dimension to assemble a sequence of such slices, as shown in Fig.~\ref{fig:Monocular depth estimation}(b). It then utilizes a Long Short-Term Memory (LSTM) network\cite{shi2015convolutional} to restore the spatial relationships among the slices, both in the short and long term. HoHoNet~\cite{sun2021hohonet} squeezes ERP features along the vertical direction and unsqueezes the compressed features for dense depth prediction, which is fast and accurate. More recently, PanelNet~\cite{yu2023panelnet} introduces to segment the ERP into panels. PanelNet employs window attention to extract local geometric information and applies a specially designed panel attention mechanism to capture global geometric information.

\noindent \textbf{Geometric Information Prior:} With large FoV, ODI provides rich and complete geometric information in the scene. Exploiting the geometric information in the ODI, \eg, edge and surface normal, can improve the performance effectively, as shown in Fig.~\ref{fig:Monocular depth estimation}(c).  Eder \etal~\cite{eder2019pano} proposed a plane-aware learning scheme that jointly predicts depth, surface normal, and boundaries. Similar to~\cite{eder2019pano}, Feng \etal~\cite{feng2020deep} proposed a framework to refine ODI depth estimation using the surface normal and uncertainty scores. Moreover, Jin \etal~\cite{jin2020geometric} demonstrated that the representations of geometric structure, \ie, corners, boundaries, and planes, can provide useful regularization for ODI depth estimation.

\noindent \textbf{Multiple Views:} Given that depth annotations for ODIs are costly, several studies have focused on multiple viewpoints to synthesize data and achieve competitive results. Zioulis~\etal~\cite{zioulis2019spherical} investigated spherical view synthesis for self-supervised ODI monocular depth estimation. In their work, upon predicting the depth map in ERP format, stereo viewpoints with vertical and horizontal baselines are generated using depth-image-based rendering, as shown in Fig.~\ref{fig:Monocular depth estimation}(d). These synthesized images are then aligned with real images sharing the same viewpoints, employing a photometric image reconstruction loss for supervision. To enhance both accuracy and stability, Yun~\etal~\cite{yun2021improving} introduced a joint learning framework that estimates ODI depth through supervised learning while concurrently estimating poses from adjacent ODI frames through self-supervised learning. Zhuang~\etal~\cite{zhuang2023spdet} proposed a transformer-based self-supervised depth prediction model, namely SPDET. In detail, SPDET first predicts a depth map from the source-view panorama, then introduces the spherical geometry feature to learn the target-view panorama from the source-view panorama, and finally achieves self-supervised depth prediction via supervising the target-view panorama generation. Similarly, Chang~\etal~\cite{chang2023depth} proposed to utilize the Neural Radiance Field (NeRF)~\cite{mildenhall2021nerf} to achieve unsupervised panoramic depth estimation. They re-projected the pixels of panoramas onto the unit sphere to get the camera rays and then supervised the generation of novel-view ODIs, based on the predicted depth maps, to achieve unsupervised depth estimation.

\noindent \textbf{Depth Completion:} Due to the scarcity of real-world sparse-to-dense panoramic depth maps, this task mainly utilizes simulation techniques to generate artificially sparse depth maps as training data. Liu~\etal~\cite{liu2022cross} proposed a representative two-stage framework to achieve panoramic depth completion. In the first stage, a spherical normalized convolution network is proposed to predict the initial dense depth maps and confidence maps from the sparse depth inputs. Then the output of the first stage is combined with corresponding ODIs to generate the final panoramic dense depth maps through a cross-modal depth completion network. Especially, BIPS~\cite{oh2021bips} proposes a GAN framework to synthesize RGB-D indoor panoramas from the limited input information about a scene captured by the camera and depth sensors in arbitrary configurations. However, BIPS ignores a large distribution gap between synthesized and real LIDAR scanners, which could be better addressed with domain adaptation techniques. Moreover, there are several methods which design depth pre-training~\cite{yan2022multi} and spherical uncertainty loss~\cite{yan2023distortion}, respectively.

\noindent \textbf{Stereophonic Depth Estimation:} Li~\etal~\cite{li2005spherical}
 used a pair of ODIs as input for estimating the depth in the scene, as the pioneering work in this field, who . They utilized the Lucas-Kanade~\cite{lucas1981iterative} method to construct the relationships between the paired ODIs. As for deep learning-based methods, Wang~\etal~\cite{wang2020360sd} extracted features from ODIs with top and bottom views and calculated the cost volume from the features with two views. Finally, the disparity is regressed from the cost volume with stacked CNNs. Lai~\etal~\cite{lai2019real} outputted both disparity and normal maps, in which the multi-task learning constrained the rationality of both types of outputs. However, as non-overlapped regions between paired ODIs are limited, this direction is scarce.

% \noindent \textbf{Discussion:} Based on the aforementioned analysis, most methods only consider indoor scenes due to two main reasons: (i) Some geometric priors are ineffective in the wild, \eg, the plane assumption; (ii) Outdoor scenes are more challenging due to the scale ambiguity in approximately infinite regions (\eg, sky), and objects in various shapes and sizes~\cite{feng2022360}.

\noindent \textbf{Discussion:} In Table~\ref{tab:deph/comparison-depth}, we evaluate the performance of monocular depth estimation methods for ODIs across two datasets. Our findings indicate that: 1) Utilizing projection types with less distortion proves to be an effective strategy for enhancing performance. Specifically, TP incurs less distortion compared to CP, resulting in TP-based methods outperforming those based on CP. 2) Incorporating an attention mechanism shows superiority over traditional CNN-based methods, primarily because attention mechanisms are expert at capturing global structural information, which is particularly advantageous for ODIs that contain abundant structural scene details. 3) The choice of network backbone is crucial. For instance, S2Net~\cite{li2023mathcal} utilizes the Swin Transformer~\cite{Liu2021SwinTH} as its backbone, yielding significantly better performance than other methods on the Matterport3D dataset.

Moreover, based on the aforementioned analysis, we think some directions are also worth exploring, except for improving the depth estimation performance. For example, as the ODI application requires extremely high spatial resolution, especially for mobile devices, \eg, VR headsets. Current methods often output depths with $512 \times 1024$ spatial resolution, which can not support immersive experience. It is interesting to investigate how to improve the computational efficiency, and simultaneously improve the output resolution.

\vspace{-5pt}
\subsubsection{ Optical Flow Estimation}
\label{sec5.3.2}
\vspace{-5pt}

Optical flow estimation analyzes pixel-level motion between two consecutive video frames, offering valuable insights for dynamic scene understanding. ODIs, with their ultra-wide field of view (FoV), provide a comprehensive perspective of the surrounding environment, making optical flow estimation based on ODIs particularly important for practical applications such as autonomous driving. However, existing optical flow methods designed for perspective images have been shown to perform poorly on ODIs~\cite{apitzsch2018cubes3d}. To address this, Xie~\etal~\cite{xie2019effective} introduced a diagnostic dataset, FlowCLEVR, to evaluate the performance of tailored convolutional filters—namely, correlation, coordinate, and deformable convolutions—for estimating optical flow in ERP format. Given the scarcity of panoramic optical flow datasets, other studies~\cite{Seidel2021OmniFlowHO},~\cite{bhandari2021revisiting},~\cite{artizzu2021omniflownet} have utilized domain adaptation strategies to leverage the wealth of perspective image datasets. Drawing inspiration from representation learning methods~\cite{Su2019KernelTN},~\cite{coors2018spherenet}, these studies exploit the geometric relationship between tangent planes and the spherical surface to re-project square convolutional filter weights learned from perspective images onto the spherical surface of ODIs. Li~\etal~\cite{li2022deep} and Yuan~\etal~\cite{yuan2021360} further addressed distortion challenges by employing less-distorted projection formats. They first applied perspective-based methods to predict optical flow in these projections and then integrated the patch-based results into ERP format. More recently, PanoFlow~\cite{shi2023panoflow} leveraged the cyclic nature of ODIs, similar to~\cite{Liao2022CylinPaintingS3}, to reduce large displacements into smaller ones, simplifying the estimation process.

\noindent\textbf{Discussion:} Despite these advancements, panoramic optical flow remains a largely underexplored area, presenting abundant opportunities for future research. Key directions include, but are not limited to, the collection of large-scale real-world datasets, the development of specialized network architectures to enhance prediction accuracy, and the integration of powerful large-scale vision models to push the boundaries of this field further.
\vspace{-5pt}

\begin{figure}[!t]
  \centering
  \includegraphics[width=\linewidth]{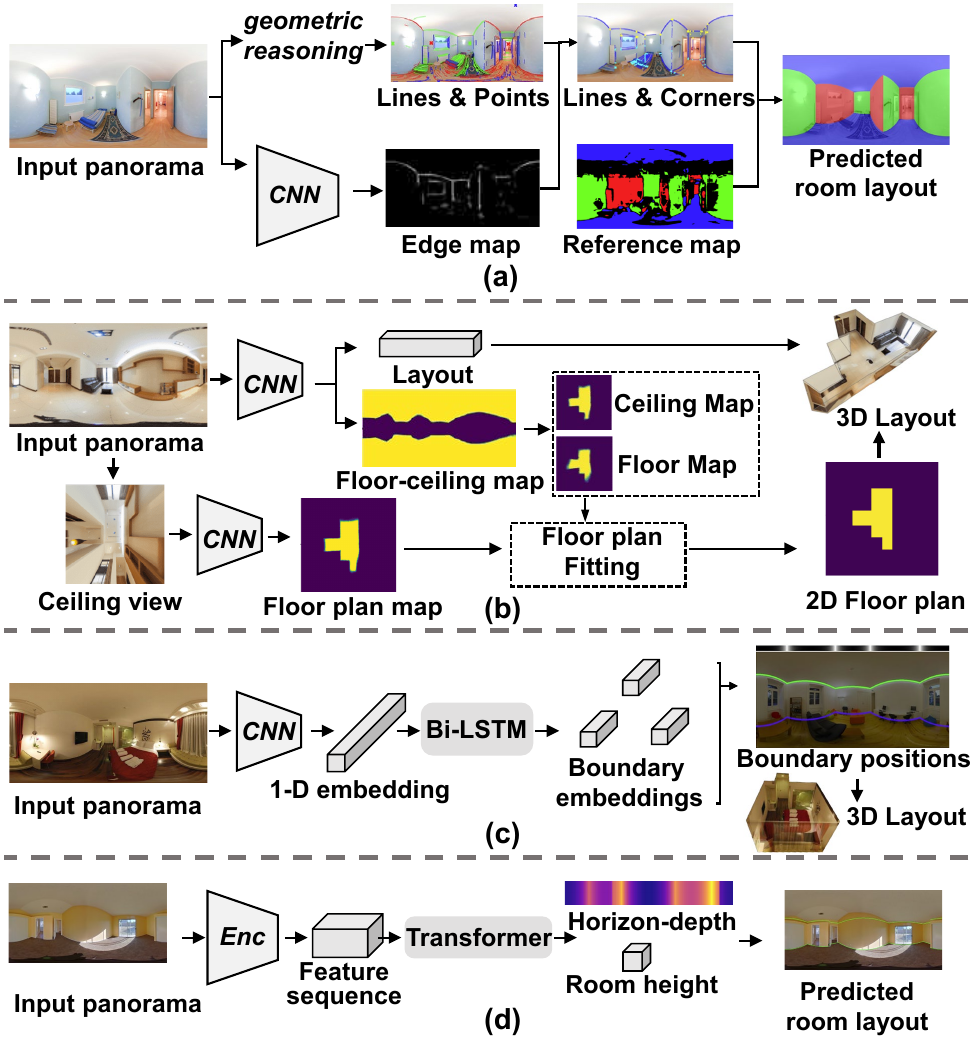}
  \vspace{-8pt}
  \caption{Representative room layout estimation methods: (a) Based on the combination of geometric reasoning and deep patterns~\cite{fernandez2018layouts}; (b) Based on panorama-view and ceiling-view inputs~\cite{Yang2019DuLaNetAD}; (c) Based on three 1D boundary embeddings~\cite{Sun2019HorizonNetLR}; (d) Based on predicting the depth of the horizon line and room height~\cite{Wang2021LED2NetM3}.} 
    \label{fig:Room Layout Estimation}
    \vspace{-10pt}
\end{figure}

\begin{table*}[!t]
    \centering
    \caption{Quantitative comparison of the methods for cuboid room layout estimation on PanoContext.
    }
    \vspace{-5pt}
    \label{tab:deph/cuboid-room-layout}
    \resizebox{0.9\linewidth}{!}{ 
    \begin{tabular}{c|c|c|c|c|c|c}
    \toprule
    Datasets&Method & Publication & Backbone & 3D IoU ($\%$) $\uparrow$ & Corner error ($\%$) $\downarrow$ & Pixel error ($\%$) $\downarrow$ \\
    \midrule
    \multirow{11}*{PanoContext~\cite{Zhang2014PanoContextAW}} 
&LayoutNet~\cite{Zou2018LayoutNetRT}&CVPR'18&-&74.48&1.06&3.34\\
&DulaNet~\cite{Yang2019DuLaNetAD}&CVPR'19&ResNet-18&77.42&-&-\\  			&CFL~\cite{FernandezLabrador2020CornersFL}&RAL'20&ResNet-50 &78.87&0.75&2.6\\  
&HorizonNet~\cite{Sun2019HorizonNetLR}&CVPR'19&ResNet-50 &82.17&0.76&2.20\\
&LayoutNet v2~\cite{Zou2021ManhattanRL}&IJCV'21&ResNet-34&85.02&0.63&1.79\\
&DuLa-Net v2~\cite{Zou2021ManhattanRL}&IJCV'21&ResNet-50&83.77&0.81&2.43\\
&SSC~\cite{Zioulis2021SingleShotCG}&-&-&83.97&0.63&1.78\\
&LED$^2$-Net~\cite{Wang2021LED2NetM3}&CVPR'21&ResNet-50&82.75&-&-\\
&LGT-Net~\cite{jiang2022lgt}&CVPR'22&ResNet-50&84.94&0.69&2.07\\
&DOP~\cite{shen2023disentangling}&CVPR'23&ResNet&85.00&0.69&2.13\\
% \midrule
% \multirow{12}*{Stanford2D3D~\cite{armeni2017joint}} 
% &LayoutNet~\cite{Zou2018LayoutNetRT}&CVPR'18&-& 76.33&1.01&2.70\\
% &DulaNet~\cite{Yang2019DuLaNetAD}&CVPR'19&ResNet-18&79.36&-&-\\ 
% &HorizonNet~\cite{Sun2019HorizonNetLR}&CVPR'19&ResNet-50 &79.79&0.71&2.39\\
% &AtlantaNet~\cite{Pintore2020AtlantaNetIT}&ECCV'20&ResNet-50 &82.43&0.70&2.25\\
% &LayoutNet v2~\cite{Zou2021ManhattanRL}&IJCV'21&ResNet-34& 84.17&0.71&2.04\\
% &DuLa-Net v2~\cite{Zou2021ManhattanRL}&IJCV'21&ResNet-50&86.60&0.67&2.48\\
% &SSC~\cite{Zioulis2021SingleShotCG}&-&-&87.80&0.51&1.62\\
% &LED$^2$-Net~\cite{Wang2021LED2NetM3}&CVPR'21&ResNet-50&83.77&-&-\\
% &Deep3DLayout~\cite{pintore2021deep3dlayout}&TOG'21&ResNet-18&89.39&-&-\\
% &LGT-Net~\cite{jiang2022lgt}&CVPR'22&ResNet-50&86.03&0.63&2.11\\
% &DOP~\cite{shen2023disentangling}&CVPR'23&ResNet&85.58&0.66&2.10\\
    \bottomrule
    \end{tabular}}
    \vspace{-15pt}
\end{table*}

\vspace{-15pt}
\subsubsection{Room Layout Estimation and reconstruction}
\vspace{-5pt}
\label{sec5.3.3}
Room layout estimation and reconstruction aims to learn the sparse 3D representation of an indoor scene. Specifically, to reduce the difficulty, rooms are often modeled as cuboids, or with either the Manhattan World~\cite{Coughlan2000TheMW} or Atlanta World~\cite{Schindler2004AtlantaWA}. Manhattan World assumes that there exist three mutually orthogonal vanishing directions in the scene, while there is one additional degree of freedom in the Atlanta World assumption. For room layout estimation and reconstruction methods on perspective images, they face two limitations: 1) the limited field of view prevents the reconstruction of the complete closed geometry of the room; 2) the ceiling does not usually appear in perspective images, which is essential for detecting the main structure of the room. Therefore, ODIs are naturally suitable for this task as they can capture the entire surroundings in a single shot.

Fernandez-Labrador~\etal~\cite{fernandez2018layouts} proposed the first deep learning-based framework for recovering room layouts from ODIs (See Fig.~\ref{fig:Room Layout Estimation}(a)). They integrated geometric reasoning (lines and vanishing points) with extracted deep patterns,~\ie edge maps, to predict structural corners. After obtaining the ceiling-wall and floor-wall corners, the authors estimated room height based on Manhattan World constraints and refined layout results using learned surface normal maps for the final output. To accurately predict edge maps of ODIs and mitigate distortions, they initially divided an ODI into 60 overlapping narrow-FoV perspective patches. Subsequently, they fused the edge maps of all patches, predicted by a pre-trained 2D edge detector, back into the ERP format. Differing from~\cite{fernandez2018layouts}, LayoutNet~\cite{Zou2018LayoutNetRT} directly learns
rooms' boundary and corner maps simultaneously through a two-branch network, consisting of one encoder and two decoders. The network takes joint ERP format inputs from ODIs and Manhattan line segments. Additionally, Manhattan line segments, serving as geometrical cues, are re-projected into the ERP format from detected line segments~\cite{von2008lsd} of overlapping perspective patches and also utilized to align the input panorama upright. Finally, following Manhattan World constraints, LayoutNet further refines the estimations and obtain the final layouts.

As geometric reasoning brings extra computation cost and inference time, some approaches propose end-to-end trainable pipelines without no additional geometric reasoning. As illustrated in Fig.~\ref{fig:Room Layout Estimation}(b), DuLa-Net~\cite{Yang2019DuLaNetAD} leverages the complementary room layout clues in two projections,~\ie, panorama-view and perspective ceiling-view to recover layout, to build a two-stream network. The floor-ceiling probability map and layout height are predicted from the panorama-view, while 2D floor plan is learned from ceiling-view. The floor-ceiling probability map and 2D floor plan are 2D binary mask maps, and the layout height is a scalar value. DuLa-Net fuses the features of two projections via the E2P conversion module (\textit{similar to the E2C module in BiFuse~\cite{wang2020bifuse}}), which transforms ERP features to perspective ceiling-view format. Finally, DuLa-Net optimizes the floor plan according to the Manhattan assumption. As the size of 2D probability maps or floor plans is $O(HW)$, HorizonNet~\cite{Sun2019HorizonNetLR} proposes a novel efficient representation specifically tailored for room layout estimation, namely three 1D vectors (the output size=$O(3\times W)$), that predicts the boundary positions of floor-wall and ceiling-wall, and the existence of wall-wall boundary (See Fig.~\ref{fig:Room Layout Estimation}(c)). Especially, to capture global information and model long-term dependencies for geometric patterns, bidirectional LSTM~\cite{medsker2001recurrent} is employed to process the feature embeddings compressed along the vertical axis. Besides efficiency, horizontal features implicitly handle the distortions of spherical images along the rows.  After obtaining the compact three representations, HorizonNet adds the Manhattan world constraint, that the ground has been aligned orthogonally to the y-axis (horizontal ground), to obtain the closed layout. This horizontal representation strategy is also employed in both HoHoNet~\cite{sun2021hohonet} and SSLayout~\cite{Tran2021SSLayout360SI}. In particular, HoHoNet proposes an efficient height compression (EHC) module to reduce the feature height to 1 and replace the Bi-LSTM in HorizonNet with a more efficient and effective multi-head
self-attention module. SSLayout is the first semi-supervised approach to use a combination of labeled
and unlabeled data for improved layout estimation. In SSLayout, HorizonNet is employed as a stochastic predictor with the dual role of being both the student and teacher and leverages the exponential moving average (EMA)~\cite{laine2017temporal} to achieve semi-supervised learning. Furthermore, LED$^2$-Net~\cite{Wang2021LED2NetM3} formulate the task of ODI-based layout estimation as predicting depth along the horizon line of a panorama (target output size = $O(W)$). Specifically, LED$^2$-Net encodes the room layout as a set of rays, called horizon-depth representation, which are equiangularly sampled from longitude on the floor-plane (Fig.~\ref{fig:Room Layout Estimation}(d)). Based on this, the 3D layout is obtained under the assumption of the gravity-aligned camera with approximately known height, and Manhattan World constraints. By contrast, LGT-Net~\cite{jiang2022lgt} proposes a transformer-based framework to predict the horizon depth and room height simultaneously. To better model the enclosed layout, the authors computed the normal and gradient errors calculated from horizon depth to constrain the planar geometry. Also predicting the room height and horizon depth with the Manhattan World assumption, Shen~\etal~\cite{shen2023disentangling} proposed two improvements: 1) producing two 1D representations from the semantics of two disentangled orthogonal planes; 2)adaptively integrating shallow and deep features with distortion awareness. The former can address the ambiguous interpretability from the compression procedure, while the latter can enhance the feature representation. 

In particular, some works follow the cuboid-shaped or Atlanta World assumption. Single-Shot Cuboids~\cite{Zioulis2021SingleShotCG} is one of the representative works for cuboid-shaped layout estimation. Without any post-processing, Single-Shot Cuboids predicts geodesic heatmaps to regress corner coordinates and constrains the geometry to a cuboid. Due to the Atlanta World assumption not requiring vertical walls to be orthogonal to other layout elements, it is more lenient than the Manhattan World assumption, while also increasing the difficulty of layout recovery. The first work based on Atlanta World assumption is AtlantaNet~\cite{Pintore2020AtlantaNetIT}, which takes the above-camera view and below-camera view of the original gravity-aligned panorama as the input rather than the whole panorama. This strategy can avoid noise from furniture and occlusions. With the estimated floor plan and ceiling plan, AtlantaNet infers the room height and recovers the 3D layout based on the Atlanta World constraints

Some approaches explore the potential to reconstruct 3D layouts from panoramas without relying on these complex structural assumptions. For instance, CFL~\cite{FernandezLabrador2020CornersFL} only considers the floor-ceiling parallelism and recovers the room layouts according to the predicted heatmaps of corners and edges. By leveraging distortion-aware spherical convolutions, CFL can generalize better to camera position variations especially when the input panoramas are not gravity-aligned. In contrast, Deep3DLayout~\cite{pintore2021deep3dlayout}adopts a different approach by building a graph convolutional network to infer the room structure as a 3D mesh. It progressively deforms a graph-encoded tessellated sphere until the final 3D layout is achieved, without explicitly predicting any layout clue. Jia~\etal~\cite{Jia20223DRL} proposed a pipeline based on HorizonNet to predict an extra surface normal feature for adaptive post-processing to reconstruct layouts of arbitrary shapes. Seg2Reg~\cite{Sun2023Seg2RegD2} adapts NeRF techniques to formulate the 2D layout representation as a density field and employs flattened volume rendering to regress 1D layout depth and generate the floor-plan polygon.

All of the methods mentioned above are based on a single input panorama. However, in reality, due to the presence of many large and complex rooms, a single panoramic image is often insufficient to accurately estimate the layout of the house. Therefore, some works use two panoramas to reconstruct the house layout. Wang~\etal~\cite{wang2022psmnet} proposed PSMNet, a joint pose-layout network to predict 2D room layout and refine a noisy 3 DOF relative camera pose simultaneously. The input of PSMNet is the two panoramas captured with different viewpoints and converted ceiling-view images using the E2P module in DuLa-Net. Notably, the key block of PSMNet for camera pose alignment is based on the two converted ceiling-view images. The geometrical clues for layout estimation are generated by the view fusion. Finally, with Atlanta World constraints, the output mask is post-processed to generate the layout polygon. Similarly, GPR-Net~\cite{su2023gpr} explores the end-to-end supervised room layout estimation from two panoramas. Compared with PSMNet, GPR-Net avoids the initial registration and presents two transformers for 1D layout horizon feature extraction and relative pose regression. 

\noindent\textbf{Discussion:} From the analysis, it can be observed that, compared to other tasks, room layout estimation based on ODIs and perspective images are quite different. ODI can provide complete ceiling and floor information, which also offers more clues to recover the complete layout. Moreover, from existing methods, we can see that most of the current methods rely on horizontal features for prediction, which imposes high requirements on the accuracy of post-processing. In contrast, Deep3DLayout proposes a promising research direction by directly recovering room layouts from the perspective of 3D meshes, enabling many existing 3D generation methods to be applied to room layout estimation. Specifically, most existing methods focus on central panoramas, while non-central panoramas allow for the recovery of the environment scale. Therefore,~\cite{berenguel2021scaled,berenguel2022atlanta} adopt layout estimation models for central panoramas to extract structural lines from non-central input panoramas and propose linear solvers to jointly obtain the room height and vertical walls location for reconstructing the scaled layout of the room. More methods targeting non-central panoramas should be developed in the future.

\vspace{-5pt}
\subsubsection{Simultaneous Localization And Mapping}
\vspace{-5pt}

Simultaneous Localization and Mapping (SLAM) leverages sequential RGB and depth data captured by a mobile agent to concurrently estimate the agent's position within its environment and reconstruct the surrounding 3D structure in real time. Compared to perspective images, omnidirectional images provide more structural information per movement, enabling more efficient and cost-effective SLAM. Caruso~\etal~\cite{Caruso2015LargescaleDS} adapted the monocular perspective SLAM pipeline, LSDSLAM~\cite{Engel2014LSDSLAMLD}, for omnidirectional cameras, and utilized a spherical coordinate system for tracking and depth estimation. In the domain of visual odometry, ROVO~\cite{Seok2019ROVORO} introduced an omnidirectional stereo rig using four fisheye cameras, achieving robust and accurate ego-motion estimation by incorporating spherical imaging characteristics and the positional parameters of multiple cameras.
Several other works~\cite{Matsuki2018OmnidirectionalDD},~\cite{Seok2020ROVINSRO}, ~\cite{Jayasuriya2020ActivePF},~\cite{Won2020OmniSLAMOL},~\cite{Wang2022LFVIOAV},~\cite{Wang2022LFVISLAMAS},~\cite{Yang2024MCOVSLAMAM},~\cite{Xie2024OmnidirectionalDS} leverage multi-camera systems to achieve omnidirectional perception. These approaches utilize each wide-FoV camera's observations for pose estimation, key-point matching, and motion tracking within constrained spaces. Subsequently, the parameters of the entire multi-camera system are utilized for omnidirectional loop-closing, resulting in a consistent global solution for localization, tracking, and mapping.

In contrast,~\cite{Huang2022360VOVO},~\cite{Kim2023CalibratingPD} focus on utilizing single 360$^\circ$ cameras to capture scenes. These approaches exploit pixel and depth information from adjacent panoramas recorded during the agent's movement to perform localization and mapping. Compared to multi-camera systems, these methods reduce alignment overhead and associated costs but face challenges in handling the increased semantic complexity inherent in 360$^\circ$ data. Generally, current omnidirectional SLAM research primarily focuses on omnidirectional perception systems constructed from multiple fisheye cameras. 
With the commercialization of 360$^\circ$ cameras, developing omnidirectional SLAM systems that leverage a single 360$^\circ$ camera for high-precision localization and mapping presents substantial practical value. For example, 360Loc~\cite{huang2024360loc} proposes to combine a single 360$^\circ$ camera and LiDAR for precise visual localization that supports the reference images with different FoVs.  Furthermore, integrating advanced 3D reconstruction techniques, such as NeRFs~\cite{mildenhall2021nerf} or 3D Gaussian splatting~\cite{Kerbl20233DGS}, into the omnidirectional SLAM pipeline offers promising avenues for future exploration.

\vspace{-10pt}
\section{Application}
\vspace{-5pt}

\subsection{AR and VR}
\vspace{-5pt}
\label{sec6.1}
With the advancement of techniques and the growing demand for interactive scenarios, AR and VR have seen rapid development in recent years. VR aims to simulate real or imaginary environments, where a participant can obtain immersive experiences and personalized content by perceiving and interacting with the environment. With the advantage of capturing the entire surrounding environment, 360 VR/AR facilitates the development of immersive experiences. \cite{kittel2020360} gives a detailed SWOT (namely strengths, weaknesses, opportunities, and threats) analysis of 360 VR to make sure that it is suitable to leverage 360 VR to develop athletes' decision-making skills. Understanding human behaviors is crucial for the application of 360 VR.~\cite{wu2020spherical} proposes a preference-aware framework for viewport prediction, and \cite{XuDWSSYG18} combined the history scan path with image contents for gaze prediction. In addition, to enhance the immersive experience, Kim~\etal~\cite{KimHJH19} proposed a novel pipeline to estimate room acoustic for plausible reproduction of spatial audio with $360^{\circ}$ cameras. Importantly, obtaining high-quality and editable panoramas can provide users with a better viewing experience with VR/AR devices. Therefore, panoramic image generation methods,~\eg,PanoDiffusion~\cite{wu2024panodiffusion}, Text2Light~\cite{Chen2022Text2Light}, are very helpful for VR/AR. Meanwhile, depth data is strongly desired in VR/AR to provide a sense of 3D. However, previous consumer-level depth sensors can only capture perspective depth maps, and panoramic depths need time-consuming stitching technologies. Therefore, accurate monocular depth estimation techniques, \eg, UniFuse~\cite{jiang2021unifuse} and PanoFormer~\cite{shen2022panoformer}, are promising.

\vspace{-15pt}
\subsection{Robot Navigation}
\vspace{-5pt}
\label{sec6.2}

For the related applications of ODI/ODV in robot navigation, it includes the telepresence system, surveillance, and DL-based optimization methods. The telepresence system aims to overcome space constraints to enable people to remotely visit and interact with each other. ODI/ODV is gaining popularity by providing a more realistic and natural scene, especially in outdoor activities with open environments~\cite{Heshmat2018GeocachingWA}.~\cite{Zhang2018A3} proposes a prototype of an ODV-based telepresence system to support more natural interactions and remote environment exploration, where real walking in the remote environment can simultaneously control the relevant movement of the robot platform. Surveillance aims to replace humans for security purposes, in which the calibration is vital for sensitive data. Accordingly, Pudics~\etal~\cite{Pudics2015SafeRN} proposed a safe navigation system tailored for obstacle detection and avoidance with a calibration design to obtain the proper distance and direction. Compared with NFoV images, ODIs can reduce the computational cost significantly by providing complete FoV in a single shot. Moreover, Ran \etal~\cite{Ran2017ConvolutionalNN} proposed a lightweight framework based on the uncalibrated $360^{\circ}$ cameras. The framework can accurately estimate the heading direction by formulating it into a series of classification tasks and avoid redundant computation by saving the calibration and correction processes. To address dark environments, \eg, underground mine, Mansouri \etal~\cite{Mansouri2019VisionbasedMN} utilized online heading rate commands to avoid collision in the tunnels and calculated scene depth information online. 

\vspace{-20pt}
\subsection{Autonomous Driving}
\vspace{-5pt}
\label{sec6.3}
It requires a full understanding of the surrounding environment, which omnidirectional vision excels at. Some works focus on setting up $360^{\circ}$ platform for autonomous driving~\cite{beltran2020towards}. Specifically,~\cite{beltran2020towards} introduced a multi-modal 360$^\circ$ perception proposal based on visual and LiDAR scanners for 3D object detection and tracking. In addition to the platform, the emergence of public omnidirectional datasets for autonomous driving is crucial for the application of DL methods. Caeser \etal~\cite{caesar2020nuscenes} were the first to introduce the relevant dataset which carries six cameras, five radars, and one LiDAR. All devices are with $360^{\circ}$ FoV. Recently, OpenMP dataset~\cite{zhang2022openmpd} is captured by six cameras and four LiDARs, which contains scenes in a complex environment, \eg, urban areas with overexposure or darkness.

\vspace{-10pt}
\section{Discussion and New Perspectives}
\vspace{-5pt}
\label{sec7}

\noindent\textbf{Cons. of Projection Formats.} ERP is the most prevalent projection format due to its wide FoV in a planar format. The main challenge for ERP is the increasing stretching distortion towards poles. Therefore, many works were proposed to design specific convolution filters against the distortion~\cite{coors2018spherenet,Su2017LearningSC}. By contrast, CP and tangent images are distortion-less projection formats by projecting a spherical surface into multiple planes. 
They are similar to the perspective images and therefore can make full use of many pre-trained models and datasets in the planar domain~\cite{li2022omnifusion}. However, CP and tangent images suffer from the challenges of higher computational cost, discrepancy, and discontinuity. 

We summarize two potential directions for utilizing CP and tangent images: (i) Redundant computational costs are resulted from large overlapping regions between projection planes. However, the pixel density varies among different sampling positions. The computation can be more efficient by allocating more resources for dense regions (\eg, equator) and fewer resources for sparse regions (\eg, poles) with reinforcement learning~\cite{Deng2021LAUNetLA}. (ii) Currently, different projection planes are often processed in parallel, which lacks global consistency. To overcome the discrepancy among different local planes, it is effective to explore an additional branch with ERP as the input~\cite{wang2020bifuse} or attention-based transformers to construct non-local dependencies~\cite{li2022omnifusion}. However, these constraints are mainly added to the feature maps, instead of the predictions. Moreover, the discrepancy can be also solved from the distribution consistency of predictions, \eg, the consistent depth range among different planes and the consistent uncertainty scores for the same edges and large gradient regions.

\noindent\textbf{Data-efficient Learning.} A challenge for DL methods is the need for large-scale datasets with high-quality annotations. However, for omnidirectional vision, constructing large-scale datasets is expensive and tedious. Therefore, it is necessary to explore more data-efficient methods. One promising direction is to transfer the knowledge learned from models trained on the labeled 2D dataset to models to be trained on the unlabeled panoramic dataset. Specifically, domain adaptation approaches can be applied to narrow the gap between perspective images and ODIs~\cite{ma2021densepass}. KD is also an effective solution by transferring learned feature information from a cumbersome perspective DNN model to a compact DNN model learning ODI data~\cite{Yang2021CapturingOC}. Finally, recent success of self-supervised methods, \eg,~\cite{yan2022multi}, demonstrates the effectiveness and necessity of pre-training models, which is still a blank for omnidirectional vision.

\noindent\textbf{Optical aberrations.}
With the growing demand for wearable devices, panoramic imaging systems face increasing requirements for lightweight and compact designs. While building minimalist panoramic imaging systems, how to correct optical aberrations and improve imaging quality remains a significant challenge. Recent efforts in computational aberration correction for panoramic imaging have focused on Panoramic Annular Lenses (PALs)\cite{powell1994panoramic}. ACI~\cite{jiang2022annular} introduced a learning-based framework that leverages physical priors and a physics-informed engine~\cite{barbastathis2019use} to correct panoramic optical aberrations. More recently, based on the Point Spread Function, Jiang~\cite{jiang2024minimalist} proposed a transformer-based architecture, trained on synthetic datasets from simulation, to achieve accurate aberration correction in minimalist panoramic imaging systems. Despite its practical importance, research on computational methods for panoramic aberration correction remains limited compared to perspective imaging. Leveraging knowledge transfer strategies to utilize the abundant data from perspective images presents a promising direction to address the scarcity of real-world panoramic datasets.

\noindent\textbf{Multi-modal Omnidirectional Vision.} It refers to the process of learning representations from different types of modalities (\eg, text-image for visual question answering, audio-visual scene understanding) and aligning them. This is a promising yet practical direction for omnidirectional vision. For instance, \cite{beltran2020towards} introduces a multi-modal perception framework based on the visual and LiDAR information for 3D object detection and tracking. However, existing works in this direction treat ODIs as the perspective images and ignore the inherent distortion in the ODIs. Future works may explore how to utilize the advantage of ODIs, \eg, complete FoV, to assist the representation of other modalities. Importantly, with the success of large-scale language models and vision models, it is desired to explore how to adapt these large-scale model on spherical imaging ODI data.

\noindent \textbf{Potential for Adversarial Attacks.} There exist few studies focusing on adversarial attacks towards omnidirectional vision models. Zhang \etal~\cite{zhang2022sp} proposed the first and representative attack approach to fool DNN models by
perturbing only one tangent image rendered from
the ODI. The proposed attack is sparse as it disturbs only a small part of the input ODI. Therefore, they further proposed a position searching method to search for the tangent point on the spherical surface. There are numerous promising yet challenging research problems in this direction, \eg, analyzing the generalization capacity of attacks among different DNN models for ODIs, white-box attacks for network architectures and training methods, and defenses against attacks. 

\noindent \textbf{Connections to Fisheye Images.} Both ODIs and fisheye images are with wide field-of-views (FoVs), but they differ in scope. ODIs capture a full 360$^\circ$ spherical view of the surrounding environment, providing a complete and continuous representation of the scene. In contrast, fisheye images capture a wide but limited FoV, typically up to 200$^\circ$. While fisheye cameras are more compact and straightforward in design, they are constrained by significant distortion and a partial view of the scene. Nowadays, fisheye images are commonly used in multi-camera systems to create surround-view systems for urban driving applications. Notably, there are several reviews and introductions~\cite{jakab2024surround,qian2022survey,singh2024overview} for fisheye images. Readers are encouraged to refer to these studies for further exploration.

% \noindent \textbf{Potential for Metaverse.} Metaverse aims to create a virtual world containing large-scale high-fidelity digital models, where users can freely create contents and obtain immersive interactive experience. Metaverse is facilitated by the AR and VR headsets, in which ODIs are favored due to the complete FoV. Therefore, a potential direction is to generate high-fidelity 2D/3D models from ODIs and simulate the real-world objects and scenes in great details. In addition, to help users obtain immersive experience, techniques that analyze and understand human behavior (\eg, gaze following, saliency prediction) can be further explored and integrated in the future.

\section{Conclusion}

\label{sec8}
In this survey, we comprehensively reviewed and analyzed the recent progress of DL methods for omnidirectional vision. We first introduced the principle of omnidirectional image acquisition, diverse projections, and datasets. We then presented a analytical explanation for distinctive representation learning and described the development of popular optimization strategies for omnidirectional vision. We further provided a hierarchical and structural taxonomy of the DL methods. For each task in the taxonomy, we summarized the current research status and compared existing algorithms, so that researchers can quickly understand the existing research directions. After constructing connections among current approaches via reviewing the practical applications, we discussed the pivotal problems to be solved and indicated promising future research opportunities. We hope this work can promote progress in the community.

\vspace{-15pt}
\section*{Data Availability Statement}
\vspace{-10pt}
Data availability is not applicable to this article as no new data were created or analyzed in this study

\bibliographystyle{plain}      % basic style, author-year citations
%\biboptions{authoryear}  

\end{document}